\documentclass{article} 
\usepackage{iclr2027_conference,times}
\iclrpreprint

\usepackage{amsmath,amsfonts,bm}

\def\eqref#1{equation~\ref{#1}}

\def\1{\bm{1}}

\DeclareMathAlphabet{\mathsfit}{\encodingdefault}{\sfdefault}{m}{sl}
\SetMathAlphabet{\mathsfit}{bold}{\encodingdefault}{\sfdefault}{bx}{n}

\usepackage{hyperref}
\usepackage{url}
\usepackage{microtype}
\usepackage{inconsolata}
\usepackage{graphicx}
\usepackage{subcaption}
\usepackage{booktabs}
\usepackage{adjustbox}
\usepackage{url}
\usepackage{minitoc}
\usepackage{float}
\usepackage{multirow}

\usepackage{amsmath}
\usepackage{amsfonts}
\usepackage{amssymb}
\usepackage{mathtools}
\usepackage{amsthm}
\usepackage{bm}
\usepackage{dsfont}
\usepackage{upgreek}

\usepackage{enumitem}
\usepackage{tikz}
\usepackage{makecell}
\usepackage{pgffor}

\usepackage{hyperref}
\usepackage[capitalize,noabbrev]{cleveref}

\theoremstyle{plain}

\theoremstyle{definition}

\theoremstyle{remark}

\usepackage{tabularx}

\usepackage{xcolor}
\newcommand{\revision}[1]{\textcolor{black}{#1}} 

\definecolor{bankvis}{HTML}{1565C0}
\definecolor{banktxt}{HTML}{2E7D32}
\definecolor{bankbind}{HTML}{C62828}
\definecolor{qacthl}{HTML}{DC3232}

\newcommand{\qfeat}[3]{%
  \par\addvspace{5pt}\noindent
  \textcolor{#1}{\rule{\linewidth}{0.8pt}}\par\nobreak\vspace{1.5pt}%
  \noindent{\footnotesize\textcolor{#1}{\textbf{Feature #2}}%
    \enspace\textbf{#3}}%
  \par\nobreak\vspace{2.5pt}}

\newcommand{\qhl}[2]{{\setlength{\fboxsep}{0.7pt}%
  \colorbox{qacthl!#1!white}{\strut #2}}}
\newcommand{\qelide}{\hspace{0.1em}\textcolor{black!45}{\dots}\hspace{0.1em}}
\newcommand{\qnum}[1]{\textcolor{black!45}{\scriptsize #1.}}
\newcommand{\qimg}[3]{
  \begin{minipage}[t]{#1}\centering
  \includegraphics[width=\linewidth]{#2}\\[0.5pt]%
  \textcolor{black!50}{\tiny #3}\end{minipage}}

\title{Reducing Hallucinations in Multimodal LLMs with Modality-Aware Sparse Autoencoders}

\author{%
\normalfont\mdseries
\makebox[\textwidth][c]{%
\begin{tabular}{c}
\rule{0pt}{2.3em}
Zhongyuan Liang$^{1,2}$ \ \ \ \
Thomas Hartvigsen$^{3}$ \ \ \ \ 
Irene Y. Chen$^{1,2}$ \ \ \ \
Ahmed Alaa$^{1,2}$
\\[0.7em]
$^{1}$UC Berkeley \ \
$^{2}$UCSF \ \
$^{3}$University of Virginia
\end{tabular}%
}%
}

\begin{document}
\doparttoc
\faketableofcontents

\maketitle

\begin{abstract}
Multimodal large language models (MLLMs) often hallucinate visual content absent from input images. Preference optimization is a common post-training approach to mitigate hallucinations, but existing preference signals often rely on LLM judges that can favor linguistic style over visual grounding. We thus introduce \emph{Modality-Aware Sparse Autoencoders} (MA-SAE), an SAE architecture that separates vision-specific, text-specific, and cross-modal features to provide a visual grounding signal for preference labeling. By isolating modality-specific patterns, such as hedging and stylistic scaffolding, MA-SAE learns interpretable cross-modal features from which we derive a grounding score measuring how well a response is supported by the image. Using its grounding score for preference labeling, MA-SAE substantially reduces hallucinations under Direct Preference Optimization (DPO), consistently outperforming LLM-as-Judge and other preference-labeling methods across three MLLM backbones and three benchmarks. The grounding signal further transfers to mDPO and on-policy preference tuning, where MA-SAE achieves strong overall performance. Further analysis shows that MA-SAE learns more coherent and better-aligned cross-modal features than existing SAE baselines, achieving the best performance across five feature-quality metrics. These results show that interpretable cross-modal features can provide a direct signal of visual grounding, offering an alternative to LLM-as-Judge supervision. Code is available at \url{https://github.com/AlaaLab/MA-SAE}.


\end{abstract}

\section{Introduction}\label{sec: intro}
Multimodal large language models (MLLMs) achieve strong performance on a wide range of vision-language tasks \citep{chen2022pali, liang2024survey, zheng2026luna, meng2026topoagent, xu2026lever}, yet they commonly hallucinate by generating visual content unsupported by the input image \citep{liu2024survey, huang2024visual, liu2026dualpathwaycircuitsobjecthallucination}. Preference optimization is a widely~used~post-training approach for mitigating hallucinations, with Direct Preference Optimization (DPO) emerging as a prominent method as it requires no separate reward model \citep{rafailov2023direct}. To efficiently scale DPO to large datasets, many works rely on LLM judges to label preference pairs \citep{li2024vlfeedback}. However, LLM judges often exhibit linguistic biases that favor verbose or stylistically preferred responses, even when they are weakly grounded in the visual input \citep{pmlr-v235-chen24h, lee2026mm}. Therefore, training on such labels risks propagating these biases into post-trained models.


We first demonstrate this linguistic bias in LLM judges through a controlled stylistic intervention on MMHal-Bench and show how this bias propagates through preference optimization to post-trained models \revision{(Section~\ref{sec:linguistic-bias})}. Starting from the same model responses, we inject hedging (e.g., ``appears to be'') and stylistic language (e.g., ``Based on the visual information in the image'') while preserving the underlying claims. Despite identical image-grounded content, the modified responses receive statistically significant score increases from all three LLM judges using the same evaluation prompts and are less frequently flagged as hallucinated (Table~\ref{tab:judge-style}). When LLM-judge~scores are used to label preference pairs, we find that preferred responses contain substantially more hedging than rejected responses. DPO training on these preferences further increases hedging in model outputs, indicating that the model learns judge-favored linguistic patterns rather than visual grounding.

\begin{figure}[t]
    \centering
    \includegraphics[width=\textwidth]{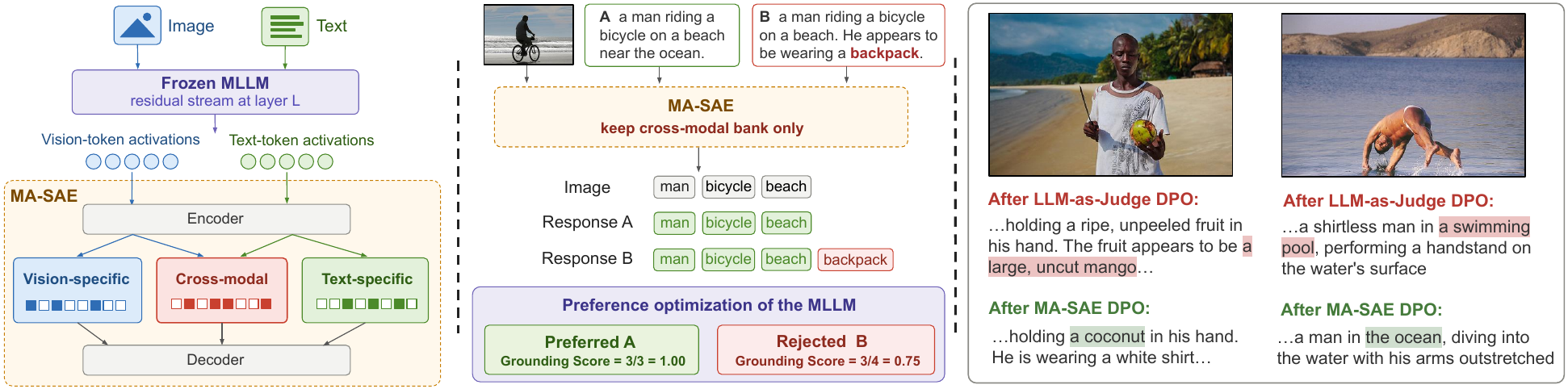}
    \caption{Left: MA-SAE decomposes MLLM representations into vision-specific, text-specific, and cross-modal feature banks. Middle: The cross-modal bank defines a grounding score used to label preference pairs for downstream optimization. Right: AMBER responses after DPO on pairs labeled by LLM-as-Judge vs. MA-SAE, where MA-SAE produces more visually grounded outputs.
    }
    \vspace{-0.2cm}
    \label{fig:overview}
\end{figure}

To obtain preference labels grounded in the image rather than in linguistic style, we propose \emph{Modality-Aware Sparse Autoencoders} (MA-SAE), an SAE that separates MLLM representations into vision-specific, text-specific, and cross-modal feature banks. This decomposition lets modality-specific banks separately capture linguistic style and visual-only structure, while encouraging the cross-modal bank to capture concepts shared across vision and text. We then derive an interpretable grounding score that measures how well a response is supported by the image, computed as the fraction of response-activated cross-modal features also activated by the image. Using this score to decide which response is preferred provides a direct visual grounding signal for preference supervision without relying on LLM judges (Figure~\ref{fig:overview}).

Across three MLLM backbones, MA-SAE learns more coherent and better-aligned cross-modal features than existing SAE approaches, achieving the best performance across all five feature quality metrics. Using the grounding score for preference labeling, MA-SAE substantially reduces hallucinations under DPO across three benchmarks, consistently outperforming LLM-as-Judge and other preference-labeling methods. These gains further generalize to mDPO \citep{wang2024mdpo} and on-policy preference tuning \citep{tajwar2024preference}, where MA-SAE continues to achieve strong performance. These results show that MA-SAE provides an interpretable cross-modal grounding signal for preference optimization, offering an alternative to supervision from LLM judges.

In summary, we make the following contributions:
\begin{enumerate}[leftmargin=*, itemsep=2pt, topsep=0pt]
    \item We demonstrate that LLM judges can reward linguistic style independently of visual grounding, and further show that this bias is reflected in LLM-judge preference labels and propagates through preference optimization to the trained model.

    \item We propose a preference-labeling method based on MA-SAE, a modality-aware SAE that separates modality-specific features capturing visual and linguistic patterns from shared cross-modal features. Its cross-modal bank yields an interpretable grounding score used to label preference pairs.

    \item We show that MA-SAE learns more coherent cross-modal features than existing SAEs across three MLLM backbones and five feature-quality metrics. Using the grounding score for preference labeling substantially reduces hallucinations across three benchmarks under DPO, with gains extending to mDPO and on-policy preference tuning.
    
\end{enumerate}

\section{Related Work}\label{sec:related}
\textbf{SAEs for MLLMs.} 
Building on the linear representation hypothesis \citep{park2023linear, liu2026ica}, SAEs decompose model activations into sparse, interpretable features \citep{bricken2023towards}. Recent work has extended SAEs to MLLMs, with the goal of learning semantically coherent features across vision and text \citep{rajaram2025line, pach2026sparse}. However, naively applying an SAE does not enforce feature sharing across modalities, and recent methods therefore introduce cross-modal alignment. VL-SAE \citep{shen2026vl} aligns paired vision and text representations using a contrastive autoencoder. GSAE \citep{kaushik2026decomposing} instead encourages shared features through group-sparse regularization. \citet{shu2026steering} align cross-modal features using a cosine-similarity objective. However, existing alignment objectives are applied over the full feature representation, which also contains modality-specific information that need not be shared across modalities. MA-SAE instead explicitly separates vision-specific, text-specific, and cross-modal features, allowing modality-specific patterns such~as linguistic style to specialize in their own banks while applying alignment only to cross-modal features.

\textbf{Biases in LLM Judges.}
LLM-as-Judge has become a widely used approach for scalable evaluation and preference supervision, but recent work has identified systematic biases in LLM-based judgments \citep{chen2024humans, zhou2024mitigating}. In particular, LLM judges can favor superficial response properties such as verbosity and fluency, even when these qualities are not directly related to correctness \citep{zhou2024mitigating, hu2025explaining}. In multimodal evaluation, LLM judges have also been shown to over-rely on linguistic cues, favoring linguistically plausible or stylistically preferred responses even when they are weakly grounded in the visual input \citep{pmlr-v235-chen24h, lee2026mm}. Such biases are particularly concerning for preference-based hallucination mitigation, where responses preferred by an LLM judge may be fluent or stylistically appealing while remaining weakly grounded in the image. Our work instead replaces LLM-as-Judge preference signals with an SAE-based grounding score that separates text-specific linguistic patterns from shared cross-modal features, yielding preferences based on visual grounding rather than linguistic style.


\textbf{Preference Construction for MLLM Hallucination Mitigation.}
Preference optimization has~become a widely used post-training approach for reducing hallucinations in MLLMs, with DPO and its variants becoming particularly prominent for directly optimizing over preference pairs without a separate reward model \citep{wang2024mdpo, xie2024v, li2026importance, li2026zooming}. Recent work has focused on how to construct and curate preference pairs using reliable scoring signals. VLFeedback \citep{li2024vlfeedback} uses LLM-as-Judge ratings to provide preference supervision, while CLIP-DPO \citep{ouali2024clip} derives preferences using CLIP image--text similarity. Other work~explores~preference data filtering and weighting rather than directly labeling preference pairs. SAE-V \citep{lou2025sae} uses vanilla SAE features to score cross-modal alignment and retains preference pairs with high alignment scores, while DA-DPO \citep{qiu2026dpo} combines CLIP and MLLM likelihood signals to weight preference pairs according to their estimated difficulty. In contrast, our work labels preference pairs using cross-modal features learned through a modality-aware SAE decomposition, providing an interpretable grounding signal while isolating linguistic patterns that may bias preference labels.

\begin{table}[t]
\centering
\small
\setlength{\tabcolsep}{4.2pt}
\renewcommand{\arraystretch}{1.08}

\begin{tabular}{lrrrrrr}
\toprule
& \multicolumn{3}{c}{LLM-judge score $\uparrow$}
& \multicolumn{3}{c}{Flagged as hallucinated (\%) $\downarrow$} \\
\cmidrule(lr){2-4}
\cmidrule(lr){5-7}
Judge
& Original & Style-injected & $\Delta$
& Original & Style-injected & $\Delta$ \\
\midrule

Gemini-3.8-Flash
& 2.38 & \textbf{2.58} & \textbf{$+0.20^{*}$}
& 56.25 & \textbf{53.12} & $-3.12$ pp \\

Claude Opus 5
& 2.19 & \textbf{2.35} & \textbf{$+0.17^{*}$}
& 62.50 & \textbf{61.46} & $-1.04$ pp \\

GPT-5.5
& 2.13 & \textbf{2.29} & \textbf{$+0.17^{*}$}
& 63.54 & \textbf{60.42} & $-3.12$ pp \\

\bottomrule
\end{tabular}

\caption{
We apply a stylistic intervention to 96 MMHal-Bench model responses by injecting stylistic language while preserving the underlying claims. Across LLM judges, modified responses receive significant score increases and are flagged as hallucinated less often. $^{*}p<0.05$ (paired Wilcoxon~test).
}
\label{tab:judge-style}
\end{table}

\section{Methodology}\label{sec:methodology}
\subsection{Preliminaries}
\label{sec:prelim}
\textbf{Multimodal Large Language Models.}
A general architecture for vision-language MLLMs consists of a vision encoder, a connector, and a language model \citep{liu2023visual, zhu2025internvl3, bai2025qwen3}. An input image is encoded into a sequence of visual tokens and projected into the language-model representation space through the connector. These visual tokens are then processed jointly with text tokens by the language model for generation. At a fixed layer, we denote the residual-stream activation of each token by $h \in \mathbb{R}^{d}$ and its modality by $m(h) \in \{V,T\}$, where $V$ and $T$ denote vision and text tokens.

\textbf{Sparse Autoencoders.}
SAEs learn sparse decompositions of model activations into features that are often more monosemantic and human-interpretable. We build on
the Top-$K$ SAE \citep{gao2025scaling}, which encodes an activation
$h \in \mathbb{R}^{d}$ as
\begin{equation}
z = \mathrm{TopK}\!\left(
\mathrm{ReLU}\!\left(W_{\mathrm{enc}}(h-b)\right)
\right),
\qquad
\hat h = W_{\mathrm{dec}} z + b,
\label{eq:topk-sae}
\end{equation}
where $W_{\mathrm{enc}} \in \mathbb{R}^{p \times d}$ and
$W_{\mathrm{dec}} \in \mathbb{R}^{d \times p}$ are the encoder and decoder
matrices, $b \in \mathbb{R}^{d}$ is the reconstruction bias, and
$z \in \mathbb{R}^{p}$ is the sparse feature representation.
$\mathrm{TopK}$ retains the $K$ largest activations and sets the remaining
entries to zero. The SAE is trained to reconstruct the original activation by
minimizing $\|\hat h-h\|_2^2$.

\subsection{Modality-aware Sparse Autoencoder}
\label{sec:masae}

To obtain a grounding signal not confounded by linguistic style, we introduce \emph{Modality-Aware Sparse Autoencoder} (MA-SAE), which extends the SAE with two modifications. First, we partition the sparse code $z$ into vision-specific, text-specific, and shared banks through hard modality gating. Second, we apply a cross-modal alignment objective only to the shared bank to encourage shared features to capture semantic concepts across vision and text.

\textbf{Modality-Partitioned Banks and Hard Gating.}
We partition the sparse code $z \in \mathbb{R}^{p}$ into three feature banks,
$z = [z_V \mid z_T \mid z_C]$, where
$z_V \in \mathbb{R}^{p_{\mathrm{vis}}}$ is vision-specific,
$z_T \in \mathbb{R}^{p_{\mathrm{text}}}$ is text-specific, and
$z_C \in \mathbb{R}^{p_{\mathrm{cross}}}$ is a shared cross-modal bank, with
$p = p_{\mathrm{vis}} + p_{\mathrm{text}} + p_{\mathrm{cross}}$.
Vision-specific features can activate only for vision tokens, text-specific features only for text tokens, while cross-modal features are accessible to both modalities.

We enforce this structure through modality-conditioned hard gating. Let $u=\mathrm{ReLU}\!\left(W_{\mathrm{enc}}(h-b)\right)
=[u_V \mid u_T \mid u_C]$ denote the pre-sparsification activations.
For a token $h$, we compute
\begin{equation}
\begin{aligned}
z_V &= \mathbb{I}[m(h)=V]\,
      \mathrm{TopK}_{K_{\mathrm{mod}}}(u_V),\\
z_T &= \mathbb{I}[m(h)=T]\,
      \mathrm{TopK}_{K_{\mathrm{mod}}}(u_T),\\
z_C &= \mathrm{TopK}_{K_{\mathrm{cross}}}(u_C),
\end{aligned}
\label{eq:hard-gating}
\end{equation}
Unlike an SAE that selects the top $K$ features globally, MA-SAE splits the sparsity budget between the two accessible banks, with
$K_{\mathrm{mod}}$ features selected from the token's modality-specific bank and $K_{\mathrm{cross}}$ from the cross-modal bank ($K = K_{\mathrm{mod}} + K_{\mathrm{cross}}$). This ensures that modality-specific features remain exclusive to their respective modality, while only cross-modal features can be shared across vision and text.

\textbf{Cross-Modal Feature Alignment.}
Multimodal SAEs can learn \emph{modality-split features}, where vision and text activate largely separate features rather than shared cross-modal concepts \citep{lee2026same}. Existing approaches encourage cross-modal sharing through group sparsity or cosine-based alignment \citep{kaushik2026decomposing, shu2026steering}. \revision{However, in our partitioned architecture, common alignment objectives can admit degenerate solutions. MSE alignment can be minimized by an inactive cross-modal bank, with modality-specific banks alone handling reconstruction, while cosine-based and KL alignment can collapse to constant cross-modal features.} We therefore introduce a Barlow Twins alignment objective \citep{zbontar2021barlow} specifically on the cross-modal bank, which aligns corresponding vision--text features while preventing collapse.

For each sample $s$, we mean-pool the cross-modal activations $z_C$ over vision and text tokens separately and standardize each feature across the batch, yielding $\bar z_{C,V}^{(s)}, \bar z_{C,T}^{(s)} \in \mathbb{R}^{p_{\mathrm{cross}}}$. We define the cross-correlation between cross-modal features $i$ and $j$ as
\begin{equation}
C_{ij}=\frac{1}{N}\sum_{s=1}^{N}
\bar z_{C,V,i}^{(s)}\bar z_{C,T,j}^{(s)}.
\end{equation}
The Barlow Twins alignment loss is
\begin{equation}
\mathcal{L}_{\mathrm{BT}}
=
\frac{1}{p_{\mathrm{cross}}}
\sum_i (1-C_{ii})^2
+
\lambda_{\mathrm{off}}
\frac{1}{p_{\mathrm{cross}}^2-p_{\mathrm{cross}}}
\sum_i \sum_{j\neq i} C_{ij}^2 .
\end{equation}

The diagonal term aligns corresponding cross-modal features across modalities and prevents collapse, \revision{while the off-diagonal term acts as a soft regularizer that reduces feature redundancy. With a small, nonzero $\lambda_{\mathrm{off}}$, this regularization encourages feature diversity without suppressing meaningful correlations between semantically related features.} The alignment loss is applied only to the cross-modal bank, leaving the modality-specific banks free to specialize.

\textbf{Training Objective.}
The full training objective combines reconstruction losses from both modalities with the Barlow Twins alignment loss,
\begin{equation}
\mathcal{L}_{\text{recon}}
= \tfrac{1}{2}\big(\mathrm{MSE}_V + \mathrm{MSE}_T\big), \ 
\text{where} \ \
\mathrm{MSE}_m = \frac{1}{|\{h : m(h)=m\}|}
\sum_{h : m(h)=m}
\|\hat h-h\|_2^2 .
\end{equation}
\begin{equation}
\mathcal{L}
= \mathcal{L}_{\text{recon}}
+ \lambda_{\text{BT}}\,\mathcal{L}_{\text{BT}} .
\end{equation}

\subsection{Grounding Score for Preference Labeling}
\label{sec:score}
We next use MA-SAE to derive preference supervision from visual grounding. We compute a~grounding score using only the cross-modal bank to measure how well a response is supported by the image. Restricting the score to cross-modal features filters out text-specific patterns in $z_T$, such as hedging or stylistic scaffolding, as well as vision-specific patterns in $z_V$ that capture visual-only structure.

Given an image and response, we run the frozen MLLM, extract the hidden states, and encode them with the frozen MA-SAE. Let $\mathcal{S}_V, \mathcal{S}_T \subseteq \{1,\ldots,p_{\mathrm{cross}}\}$ denote the sets of cross-modal features in $z_C$ activated by at least one vision or response token, respectively. We define the grounding score as
\begin{equation}
\mathrm{Ground}(\mathcal{S}_V,\mathcal{S}_T)
=
\frac{|\mathcal{S}_V \cap \mathcal{S}_T|}{|\mathcal{S}_T|},
\label{eq:grounding}
\end{equation}
which measures the fraction of response-activated cross-modal features also supported by the image.

\textbf{Preference Labeling.} We use the grounding score to label preference pairs. Given two candidate responses for the same input, the response with the higher grounding score is treated as preferred and the lower as rejected. The resulting pairs are then used for preference optimization.

\section{Experiments}\label{sec: experiments}
\subsection{Experimental Setup}
\label{sec:setup}
\textbf{Models and Compute.}
We use three MLLMs across two model families and multiple model scales: LLaVA-1.5-7B, LLaVA-1.5-13B \citep{liu2024improved}, and InternVL3-8B \citep{zhu2025internvl3}. SAE training uses a single GPU, and preference optimization uses four GPUs with DeepSpeed ZeRO-1. All experiments are conducted on NVIDIA A100 GPUs.

\textbf{MA-SAE Training.} 
We train one MA-SAE for each MLLM using residual-stream activations from the midpoint layer: layer $16$ of $32$ for LLaVA-1.5-7B, layer $20$ of $40$ for LLaVA-1.5-13B, and layer $14$ of $28$ for InternVL3-8B. We focus on intermediate layers following prior work suggesting that they capture richer semantic representations than earlier or final layers \citep{ma2026sparse}. We train the SAE on LLaVA-Instruct-150K \citep{liu2023visual}, a multimodal instruction-following dataset of image--text conversations, holding out $5\%$ of the data for evaluation. Full architecture and training hyperparameters, including bank sizes, sparsity level, learning rate, $\lambda_{\mathrm{BT}}$, are provided in Appendix~\ref{appendix:masae-training}.

\textbf{Preference Optimization.}
For the second-stage preference optimization, we construct preference pairs from VLFeedback \citep{li2024vlfeedback}, a large-scale multimodal preference dataset with responses generated by a pool of 12 MLLMs. To isolate the effect of preference labeling, all methods use the same candidate-response pools and training configuration, differing only in the labeling rule. We perform preference optimization with LoRA \citep{hu2021lora} on the language model, while keeping the vision tower and connector frozen. Full preference-pair construction details and training hyperparameters, including learning rate, batch size, and DPO $\beta$, are provided in Appendix~\ref{appendix:dpo-training}.

\subsection{Linguistic Bias in LLM-as-Judge Preference Supervision}
\label{sec:linguistic-bias}

\revision{We first demonstrate linguistic bias in LLM judges through a controlled stylistic intervention on 96 MMHal-Bench responses generated by the untuned LLaVA-1.5-7B. Starting from the same model responses, we inject hedging (e.g., changing ``there is X'' to ``there appears to be X'') and prepend stylistic language (e.g., ``Based on the visual information in the image'') while preserving the underlying claims (full rules and examples in Appendix~\ref{app:style-injection}). We then score the original and modified responses with three LLM judges using the same evaluation prompts. As shown in Table~\ref{tab:judge-style}, despite containing the same image-grounded claims, the modified responses receive statistically significant score increases from all three LLM judges and are less frequently flagged as hallucinated.}

\revision{We next examine whether this linguistic bias is reflected in preference pairs labeled by LLM-as-Judge and whether it can propagate to post-trained models through preference tuning. Using VLFeedback preference pairs labeled by LLM judges, we find that preferred responses contain 1.48$\times$ more hedging expressions than rejected responses (e.g., ``suggesting'': 1.71$\times$; ``appears to be'': 1.85$\times$). After DPO preference tuning on these pairs, LLaVA-1.5-7B generates responses with 2.3$\times$ as many hedging occurrences on average as the untuned model (e.g., ``seems to be'': 4.1$\times$; ``indicating'': 3.7$\times$). These findings show that judge-favored linguistic patterns are encoded in the preference labels and learned by the post-trained model. Full results are provided in Appendix~\ref{app:linguistic-bias}.}

\subsection{Evaluating MA-SAE Representations}
\label{sec:masae-eval}
\vspace{-0.1cm}
Before applying MA-SAE to preference optimization, we first evaluate the quality of its learned representations. By separating modality-specific and shared features, we hypothesize that MA-SAE learns more coherent and better-aligned cross-modal representations. We begin with qualitative examples examining the distinct roles of features learned by the three banks, and then quantitatively compare MA-SAE against existing SAE approaches using established feature-quality metrics.

\begin{figure}[t]
\centering
\includegraphics[width=0.95\textwidth]{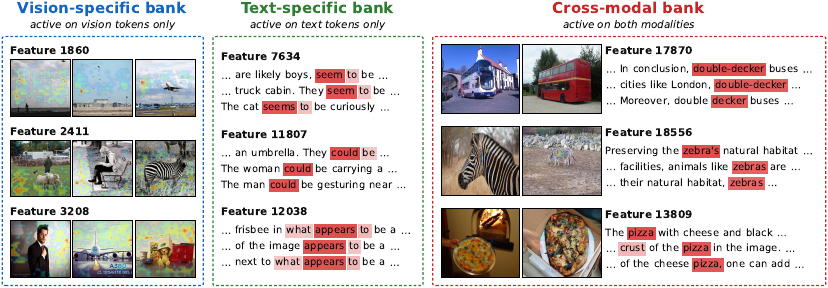}
\caption{Examples of MA-SAE features. Highlighted regions/tokens reflect high activations. Vision bank features capture visual-only structure absent from the caption, text bank features capture linguistic patterns such as hedging, and cross-modal features encode shared concepts across modalities.}
\label{fig:banks}
\vspace{-0.1cm}
\end{figure}

\textbf{Qualitative Analysis.} Figure~\ref{fig:banks} provides examples illustrating the distinct roles of features learned by the three MA-SAE banks. Vision-specific features activate on visual content not represented in the paired text, often highlighting background or scene-level regions. Text-specific features capture linguistic patterns such as hedging and inferential expressions (e.g., ``seems to''), while cross-modal features encode shared concepts across modalities (e.g., zebra images and text mentions of zebras). These examples qualitatively support the intended decomposition between modality-specific and cross-modal information. Additional examples from all three banks are provided in Appendix~\ref{appendix: examples}.

\vspace{-0.2cm}
\begin{table}[t]
\centering
\small
\setlength{\tabcolsep}{6pt}
\renewcommand{\arraystretch}{1.0}
\begin{tabular}{lccccc}
\toprule
Method
& MMS$_{v\text{-}t}\uparrow$
& MMS$_{v\text{-}v}\uparrow$
& MMS$_{t\text{-}t}\uparrow$
& Intra-Sim.$\uparrow$
& Inter-Sim.$\downarrow$ \\
\midrule

\multicolumn{6}{l}{\textit{LLaVA-1.5-7B}} \\
\quad Vanilla SAE \citep{rajaram2025line}
& 0.1599 & 0.4670 & 0.6180 & 0.2912 & 0.2656 \\
\quad VL-SAE \citep{shen2026vl}
& 0.1596 & 0.5105 & 0.5984 & 0.2735 & 0.2696 \\
\quad GSAE \citep{kaushik2026decomposing}
& 0.1723 & 0.5066 & 0.6150 & 0.2974 & 0.2741 \\
\quad \textbf{MA-SAE (ours)}
& \textbf{0.2386} & \textbf{0.6509} & \textbf{0.6960}
& \textbf{0.3513} & \textbf{0.2313} \\

\midrule
\multicolumn{6}{l}{\textit{LLaVA-1.5-13B}} \\
\quad Vanilla SAE \citep{rajaram2025line}
& 0.1589 & 0.4586 & 0.6172 & 0.2934 & 0.2638 \\
\quad VL-SAE \citep{shen2026vl}
& 0.1463 & 0.5170 & 0.5158 & 0.2770 & 0.2589 \\
\quad GSAE \citep{kaushik2026decomposing}
& 0.1731 & 0.5044 & 0.6162 & 0.2974 & 0.2709 \\
\quad \textbf{MA-SAE (ours)}
& \textbf{0.2407} & \textbf{0.6522} & \textbf{0.6982}
& \textbf{0.3526} & \textbf{0.2318} \\

\midrule
\multicolumn{6}{l}{\textit{InternVL3-8B}} \\
\quad Vanilla SAE \citep{rajaram2025line}
& 0.1843 & 0.5319 & 0.6310 & 0.3043 & 0.2705 \\
\quad VL-SAE \citep{shen2026vl}
& 0.1364 & 0.5210 & 0.5001 & 0.2636 & 0.2513 \\
\quad GSAE \citep{kaushik2026decomposing}
& 0.1820 & 0.5260 & 0.6262 & 0.3038 & 0.2708 \\
\quad \textbf{MA-SAE (ours)}
& \textbf{0.2387} & \textbf{0.6496} & \textbf{0.6950}
& \textbf{0.3509} & \textbf{0.2303} \\

\bottomrule
\end{tabular}
\caption{
Comparison of SAEs on feature-quality metrics across three MLLMs. Higher MMS and~Intra-Similarity indicate greater semantic coherence, while lower Inter-Similarity indicates better feature separation. MA-SAE consistently achieves the best performance across all metrics and models.
}
\vspace{-0.4cm}
\label{tab:monosemanticity}
\end{table}

We next quantitatively compare MA-SAE cross-modal features against existing MLLM SAE~baselines using established feature-quality metrics. We first introduce the baselines and evaluation~metrics.

\textbf{SAE Baselines.}
We compare MA-SAE against three SAE architectures for MLLMs:
(a) \emph{Vanilla SAE} \citep{rajaram2025line}, which applies a vanilla SAE to vision and text activations without explicit cross-modal alignment;
(b) \emph{VL-SAE} \citep{shen2026vl}, which aligns paired vision and text representations with a contrastive autoencoder, then preserves this similarity with a distance-based encoder;
(c) \emph{GSAE} \citep{kaushik2026decomposing}, which promotes cross-modal
feature sharing through group-sparse regularization, encouraging paired
modalities to activate a common subset of features.

\textbf{Evaluation Metrics.}
A desirable MLLM SAE should learn semantically coherent shared features across modalities, where each feature captures a consistent concept while remaining distinct from others. Following \citet{shen2026vl}, we report Intra-Similarity and Inter-Similarity. Intra-Similarity measures the cosine similarity between the mean-pooled CLIP embeddings of a feature's top-activating vision and text examples, while Inter-Similarity measures the corresponding cross-feature similarity. We additionally report the multimodal monosemanticity score (MMS) \citep{kaushik2026decomposing}, which weights semantic similarity by co-activation strength and penalizes modality-split features. We report MMS$_{v\text{-}t}$ for cross-modal coherence and MMS$_{v\text{-}v}$ and MMS$_{t\text{-}t}$ for within-modality coherence. Higher MMS and Intra-Similarity and lower Inter-Similarity indicate better feature quality. For MA-SAE, all metrics are computed over the cross-modal bank. Detailed definitions and additional matched-budget comparisons are provided in Appendices~\ref{sec:app-metrics} and~\ref{app:coverage}.

\textbf{Quantitative Results.} Table~\ref{tab:monosemanticity} shows the quantitative comparison of MA-SAE against existing SAE baselines. Across all five feature-quality metrics and all three MLLM backbones, MA-SAE consistently achieves the best performance. This suggests that separating modality-specific information from the shared feature space leads to more coherent and better-aligned cross-modal representations. Appendix~\ref{app:coverage} further reports bimodal MMS and a matched-budget comparison, where MA-SAE continues to perform best across all three backbones.

\textbf{Ablation Study.} We ablate the MA-SAE architecture in Appendix~\ref{app:masae-ablation}. Results show that adding modality-partitioned banks improves feature quality over the vanilla SAE, with the strongest performance achieved when combined with Barlow Twins alignment in the full MA-SAE.

\subsection{Preference Optimization with MA-SAE}
\label{sec:pref-opt}
We next apply MA-SAE to preference optimization using the grounding score. We first analyze the grounding signal directly, then evaluate its effectiveness when used to label preference pairs.


\textbf{Grounding Signal Analysis.} We begin by examining whether the MA-SAE grounding signal dis-
\vspace{-0.16in}
\begin{figure}[H]
\begin{minipage}{0.55\textwidth}
tinguishes image-grounded objects from hallucinated ones as desired. Figure~\ref{fig:grounding-qualitative} illustrates how cross-modal features quantify image support for individual object mentions. For each object mention, image support measures the fraction of its cross-modal features that are also activated by vision tokens. Grounded objects such as ``boys'' and ``surfboard'' receive strong image support, whereas the hallucinated ``bottle'' receives substantially lower support. Additional qualitative examples are provided in Appendix~\ref{app:grounding-qualitative}. We then evaluate this signal at scale on 500 MSCOCO samples. We group object mentions as grounded or hallucinated based on the image ground-truth annotations and average image support wi-
\end{minipage}
\enspace 
\begin{minipage}{0.43\linewidth}
\centering
\includegraphics[width=2.4in]{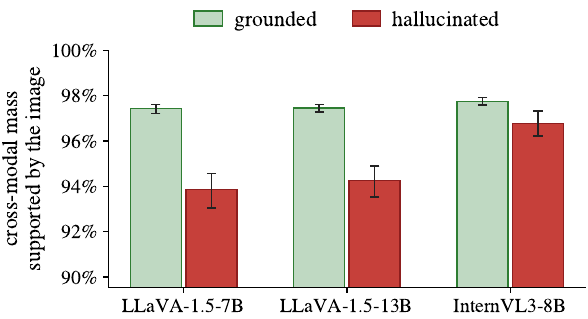}
\vspace{-.15in}
\caption{
Image support for grounded and hallucinated object mentions across three MLLM backbones.
}
\label{fig:grounding-quantitative}
\end{minipage}
\end{figure}
\vspace{-.295in}
thin each group. As shown in Figure~\ref{fig:grounding-quantitative}, grounded mentions receive significantly higher image support across all three backbones (Mann--Whitney U test, $p < 10^{-3}$ for all backbones), demonstrating that the MA-SAE grounding signal effectively separates grounded from hallucinated content.

\begin{figure}[t]
    \centering
    \includegraphics[width=0.95\linewidth]{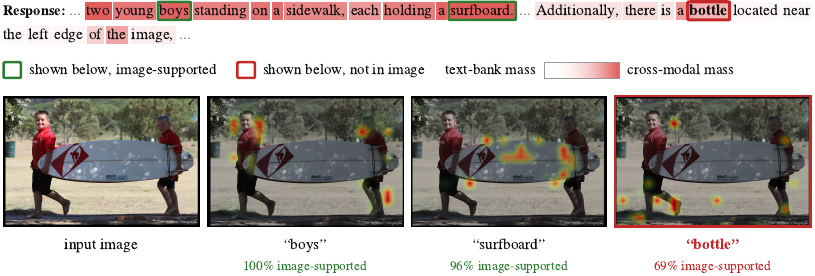}
    \vspace{-0.05in}
    \caption{
    For each object mention, we measure the fraction of cross-modal features also activated by vision tokens and visualize the corresponding image regions. Grounded objects such as ``boys'' and ``surfboard'' receive strong image support, whereas the hallucinated ``bottle'' receives lower support.
    }
    \vspace{-0.15in}
    \label{fig:grounding-qualitative}
\end{figure}

We next evaluate preference optimization using the MA-SAE grounding score to label preference pairs. We begin by introducing the preference-labeling baselines and evaluation metrics.

\textbf{Preference Labeling Baselines.}
We compare MA-SAE against four baselines for labeling preference pairs:
(a) \emph{LLM-as-Judge} \citep{li2024vlfeedback}, which uses the GPT-4V ratings provided by VLFeedback as the preference score;
(b) \emph{CLIP-DPO} \citep{ouali2024clip}, which scores each response by the cosine similarity between its CLIP text and image embeddings;
(c) \emph{SAE-V} \citep{lou2025sae}, which uses vanilla SAE features and scores responses using cosine-based weights from their top-activating vision and text examples; and
(d) \emph{DA-DPO} \citep{qiu2026dpo}, which combines the CLIP-based score with response log-probabilities from a pretrained reference MLLM.

\begin{table}[!b]
\centering
\scriptsize
\setlength{\tabcolsep}{4.5pt}
\renewcommand{\arraystretch}{1.03}

\begin{tabular}{lrrrrrrrr}
\toprule
Method
& \multicolumn{4}{c}{AMBER}
& \multicolumn{2}{c}{Object HalBench}
& \multicolumn{2}{c}{MSCOCO} \\
\cmidrule(lr){2-5}
\cmidrule(lr){6-7}
\cmidrule(lr){8-9}
& CHAIR$\downarrow$
& Cover$\uparrow$
& HalRate$\downarrow$
& Cog$\downarrow$
& CHAIR$_s\downarrow$
& CHAIR$_i\downarrow$
& CHAIR$_s\downarrow$
& CHAIR$_i\downarrow$ \\
\midrule

\multicolumn{9}{l}{\textbf{DPO \citep{rafailov2023direct}}} \\
\midrule

\multicolumn{9}{l}{\textit{LLaVA-1.5-7B}} \\
\quad \textcolor{gray}{Untuned}
& \textcolor{gray}{7.7}
& \textcolor{gray}{51.8}
& \textcolor{gray}{35.6}
& \textcolor{gray}{4.4}
& \textcolor{gray}{52.67}
& \textcolor{gray}{13.92}
& \textcolor{gray}{60.80}
& \textcolor{gray}{19.63} \\
\quad LLM-as-Judge
& 5.8 & 54.3 & 32.7 & 2.4
& 43.67 & 10.72 & 52.00 & 15.09 \\
\quad CLIP-DPO \citep{ouali2024clip}
& 5.3 & \textbf{55.3} & 30.4 & 2.8
& 42.33 & 9.99 & 50.60 & 14.42 \\
\quad SAE-V \citep{lou2025sae}
& 7.6 & 54.1 & 39.5 & 4.3
& 59.33 & 16.32 & 65.80 & 21.03 \\
\quad DA-DPO \citep{qiu2026dpo}
& 5.5 & 53.5 & 31.9 & 2.9
& 51.33 & 12.62 & 56.00 & 16.00 \\
\quad \textbf{MA-SAE (ours)}
& \textbf{4.3} & 48.0 & \textbf{17.2} & \textbf{1.5}
& \textbf{18.67} & \textbf{6.38}
& \textbf{29.00} & \textbf{10.88} \\
\midrule

\multicolumn{9}{l}{\textit{LLaVA-1.5-13B}} \\
\quad \textcolor{gray}{Untuned}
& \textcolor{gray}{6.6}
& \textcolor{gray}{51.8}
& \textcolor{gray}{30.8}
& \textcolor{gray}{3.3}
& \textcolor{gray}{50.67}
& \textcolor{gray}{13.98}
& \textcolor{gray}{58.20}
& \textcolor{gray}{19.01} \\
\quad LLM-as-Judge
& 6.1 & 54.8 & 33.6 & 2.7
& 41.33 & 10.68 & 49.60 & 15.16 \\
\quad CLIP-DPO \citep{ouali2024clip}
& 5.7 & \textbf{55.5} & 31.5 & 2.5
& 45.67 & 10.51 & 51.40 & 14.37 \\
\quad SAE-V \citep{lou2025sae}
& 6.8 & 53.6 & 34.4 & 3.5
& 51.00 & 13.79 & 57.00 & 18.57 \\
\quad DA-DPO \citep{qiu2026dpo}
& 5.4 & 53.6 & 29.8 & 2.6
& 46.67 & 10.74 & 54.20 & 15.13 \\
\quad \textbf{MA-SAE (ours)}
& \textbf{4.1} & 49.6 & \textbf{17.9} & \textbf{1.3}
& \textbf{26.67} & \textbf{7.75}
& \textbf{36.60} & \textbf{12.74} \\

\midrule

\multicolumn{9}{l}{\textit{InternVL3-8B}} \\
\quad \textcolor{gray}{Untuned}
& \textcolor{gray}{7.5}
& \textcolor{gray}{71.2}
& \textcolor{gray}{62.4}
& \textcolor{gray}{5.4}
& \textcolor{gray}{27.00}
& \textcolor{gray}{7.60}
& \textcolor{gray}{34.40}
& \textcolor{gray}{11.23} \\
\quad LLM-as-Judge
& 7.9 & 71.2 & 63.2 & 5.7
& 30.33 & 7.63 & 38.00 & 11.66 \\
\quad CLIP-DPO \citep{ouali2024clip}
& 7.7 & 72.3 & 63.3 & 6.2
& 26.67 & 7.71 & 36.00 & 10.84 \\
\quad SAE-V \citep{lou2025sae}
& 7.2 & 70.2 & 59.1 & 5.2
& 28.67 & 7.46 & 37.20 & 11.51 \\
\quad DA-DPO \citep{qiu2026dpo}
& 7.4 & 71.1 & 61.5 & 5.5
& 30.33 & 7.53 & 37.60 & 11.05 \\
\quad \textbf{MA-SAE (ours)}
& \textbf{6.9} & \textbf{72.5} & \textbf{57.4} & \textbf{5.0}
& \textbf{21.00} & \textbf{6.36}
& \textbf{29.80} & \textbf{10.25} \\

\bottomrule
\end{tabular}

\caption{
Off-policy preference tuning with DPO. We compare MA-SAE with preference-labeling baselines. MA-SAE achieves the strongest overall performance across metrics and backbones.
}
\label{tab:hallu-dpo}
\end{table}

\textbf{Benchmarks and Metrics.}
We evaluate generative hallucination on three widely used benchmarks \citep{wang2024mdpo, ouali2024clip, lu2025dama, qiu2026dpo}: AMBER \citep{wang2023amber}, Object HalBench \citep{rohrbach2018object}, and MSCOCO \citep{hua2025steering}. AMBER evaluates object hallucination using CHAIR, object coverage (Cover), response-level hallucination rate (HalRate), and cognitive hallucination (Cog). Object HalBench and MSCOCO report sentence-level CHAIR$_s$ and instance-level CHAIR$_i$. CHAIR$_s$ measures the fraction of generated responses containing at least one hallucinated object, and CHAIR$_i$ measures the fraction of object mentions that are hallucinated. Lower CHAIR, HalRate, and Cog indicate fewer hallucinations, while higher Cover indicates greater coverage of image content. We intentionally exclude MMHal-Bench \citep{sun2024aligning}, whose evaluation relies on an LLM judge, to avoid the stylistic biases demonstrated in Table~\ref{tab:judge-style}.



\textbf{Off-policy DPO Results.}
Table~\ref{tab:hallu-dpo} reports performance under DPO using off-policy VLFeedback. MA-SAE achieves the strongest overall performance across all three MLLM backbones and hallucination benchmarks, consistently reducing hallucinations relative to LLM-as-Judge and other baselines. These gains are consistent across backbones, suggesting that the MA-SAE grounding score provides a more reliable signal for preference labeling. \revision{To further investigate whether these improvements reflect better visual grounding rather than reduced object coverage, we conduct additional length-controlled preference-labeling and matched-coverage evaluations (Appendix~\ref{app:length-coverage}). MA-SAE maintains its advantage in both evaluations.} Figure~\ref{fig:overview} (right) shows AMBER responses after DPO on pairs labeled by LLM-as-Judge vs. MA-SAE, with additional examples in Appendix~\ref{app:qualitative-dpo}.

We additionally examine whether preferences labeled by MA-SAE exhibit the hedging bias observed for LLM-derived preferences. As shown in Appendix~\ref{app:masae-preference-hedging}, preferred and rejected responses labeled by MA-SAE exhibit nearly identical levels of hedging, indicating that MA-SAE preference labeling does not exhibit the same stylistic bias.

\textbf{Generalization to mDPO.}
Because MA-SAE labels preference pairs independently of the optimization objective, it can be paired with different preference-learning methods. We next evaluate this generality with mDPO \citep{wang2024mdpo}, a multimodal DPO variant that augments response preferences with contrastive image preferences. As shown in Table~\ref{tab:hallu-mdpo}, MA-SAE remains competitive across metrics and backbones, suggesting that its benefits extend beyond the standard DPO objective.

\begin{table}[h]
\centering
\scriptsize
\setlength{\tabcolsep}{4.5pt}
\renewcommand{\arraystretch}{1.03}

\begin{tabular}{lrrrrrrrr}
\toprule
Method
& \multicolumn{4}{c}{AMBER}
& \multicolumn{2}{c}{Object HalBench}
& \multicolumn{2}{c}{MSCOCO} \\
\cmidrule(lr){2-5}
\cmidrule(lr){6-7}
\cmidrule(lr){8-9}
& CHAIR$\downarrow$
& Cover$\uparrow$
& HalRate$\downarrow$
& Cog$\downarrow$
& CHAIR$_s\downarrow$
& CHAIR$_i\downarrow$
& CHAIR$_s\downarrow$
& CHAIR$_i\downarrow$ \\
\midrule

\multicolumn{9}{l}{\textbf{mDPO \citep{wang2024mdpo}}} \\
\midrule

\multicolumn{9}{l}{\textit{LLaVA-1.5-7B}} \\
\quad LLM-as-Judge
& 4.2 & 52.4 & 23.7 & 2.0
& 37.33 & 10.81 & 45.20 & 14.26 \\
\quad CLIP-DPO \citep{ouali2024clip}
& 4.6 & 52.4 & 25.6 & 2.2
& 37.67 & 10.29 & 45.20 & 14.25 \\
\quad SAE-V \citep{lou2025sae}
& 4.7 & \textbf{53.4} & 24.9 & 2.0
& 36.33 & 9.06 & 44.20 & 13.10 \\
\quad DA-DPO \citep{qiu2026dpo}
& 4.9 & 51.6 & 27.2 & 2.3
& 44.00 & 11.73 & 50.00 & 15.33 \\
\quad \textbf{MA-SAE (ours)}
& \textbf{3.8} & 51.1 & \textbf{19.4} & \textbf{1.7}
& \textbf{30.67} & \textbf{8.76}
& \textbf{37.60} & \textbf{12.53} \\

\midrule

\multicolumn{9}{l}{\textit{LLaVA-1.5-13B}} \\
\quad LLM-as-Judge
& \textbf{3.7} & 52.1 & 20.2 & \textbf{1.5}
& 35.33 & 9.49 & 42.20 & 13.58 \\
\quad CLIP-DPO \citep{ouali2024clip}
& 4.0 & 52.2 & 21.5 & 1.7
& 33.33 & 9.48 & 43.00 & 13.92 \\
\quad SAE-V \citep{lou2025sae}
& 4.4 & \textbf{52.5} & 23.9 & 1.8
& 36.67 & 9.21 & 46.20 & 13.27 \\
\quad DA-DPO \citep{qiu2026dpo}
& \textbf{3.7} & 50.2 & 20.0 & 1.7
& 39.33 & 10.90 & 46.60 & 14.80 \\
\quad \textbf{MA-SAE (ours)}
& \textbf{3.7} & 49.7 & \textbf{18.6} & \textbf{1.5}
& \textbf{27.00} & \textbf{7.60}
& \textbf{37.80} & \textbf{12.17} \\

\midrule

\multicolumn{9}{l}{\textit{InternVL3-8B}} \\
\quad LLM-as-Judge
& 7.9 & 72.0 & 63.4 & 6.3
& 33.00 & 7.81 & 40.00 & 11.99 \\
\quad CLIP-DPO \citep{ouali2024clip}
& 7.5 & 72.0 & 60.7 & 5.7
& 28.67 & 7.71 & 38.20 & 11.90 \\
\quad SAE-V \citep{lou2025sae}
& 7.5 & 70.9 & 59.3 & \textbf{5.6}
& 29.67 & 7.87 & 38.20 & 11.79 \\
\quad DA-DPO \citep{qiu2026dpo}
& 7.7 & 71.7 & 64.0 & 6.1
& 33.67 & 8.03 & 40.80 & 11.96 \\
\quad \textbf{MA-SAE (ours)}
& \textbf{7.1} & \textbf{72.5} & \textbf{57.1} & \textbf{5.6}
& \textbf{28.33} & \textbf{7.11}
& \textbf{35.20} & \textbf{11.04} \\

\bottomrule
\end{tabular}

\caption{
Generalization to mDPO. MA-SAE remains competitive with preference-labeling baselines across three MLLM backbones, showing that its benefits extend beyond standard DPO.
}
\label{tab:hallu-mdpo}
\end{table}

\textbf{Generalization to On-policy Preference Tuning.}
Experiments above use the off-policy VLFeedback dataset, where candidate responses are generated by a fixed pool of 12 MLLMs. Recent work suggests that preference optimization can benefit from on-policy data, where candidates are generated by the model being optimized \citep{tajwar2024preference, yang2025mitigating, jiang2026rocktokens}. We therefore evaluate an on-policy setting where candidate responses are sampled from the current policy at temperature 1 and then paired and labeled using the corresponding scoring method. As shown in Table~\ref{tab:hallu-onpolicy}, MA-SAE consistently outperforms LLM-as-Judge across all three backbones, demonstrating that its benefits extend beyond fixed off-policy preference data to the on-policy setting. \revision{Notably, MA-SAE achieves both higher object coverage and lower hallucination rates across all three backbones under on-policy tuning, in contrast to the coverage reductions observed for the off-policy LLaVA models. We further analyze this difference in Appendix~\ref{app:policy-length}, showing that on-policy preference pairs have smaller response-length differences, reducing length-related confounding and providing evidence that MA-SAE reduces hallucinations without sacrificing object coverage.}

\begin{table}[h]
\centering
\scriptsize
\setlength{\tabcolsep}{5pt}
\renewcommand{\arraystretch}{1.03}

\begin{tabular}{lrrrrrrrr}
\toprule
Method
& \multicolumn{4}{c}{AMBER}
& \multicolumn{2}{c}{Object HalBench}
& \multicolumn{2}{c}{MSCOCO} \\
\cmidrule(lr){2-5}
\cmidrule(lr){6-7}
\cmidrule(lr){8-9}
& CHAIR$\downarrow$
& Cover$\uparrow$
& HalRate$\downarrow$
& Cog$\downarrow$
& CHAIR$_s\downarrow$
& CHAIR$_i\downarrow$
& CHAIR$_s\downarrow$
& CHAIR$_i\downarrow$ \\
\midrule

\multicolumn{9}{l}{\textbf{On-policy DPO \citep{tajwar2024preference}}} \\
\midrule

\multicolumn{9}{l}{\textit{LLaVA-1.5-7B}} \\
\quad LLM-as-Judge
& 5.5 & 50.8 & 28.7 & 2.7
& 42.33 & 10.80 & 49.80 & 15.26 \\
\quad \textbf{MA-SAE (ours)}
& \textbf{5.1} & \textbf{51.0} & \textbf{23.8} & \textbf{1.9}
& \textbf{23.00} & \textbf{7.11}
& \textbf{33.40} & \textbf{11.93} \\

\midrule

\multicolumn{9}{l}{\textit{LLaVA-1.5-13B}} \\
\quad LLM-as-Judge
& 4.9 & 50.8 & 23.0 & 2.1
& 37.00 & 9.92 & 45.60 & 14.84 \\
\quad \textbf{MA-SAE (ours)}
& \textbf{4.5} & \textbf{52.4} & \textbf{20.9} & \textbf{1.5}
& \textbf{25.33} & \textbf{6.91}
& \textbf{34.80} & \textbf{11.23} \\

\midrule

\multicolumn{9}{l}{\textit{InternVL3-8B}} \\
\quad LLM-as-Judge
& 7.3 & 70.8 & 60.6 & 5.2
& 29.33 & 8.17 & 37.40 & 11.99 \\
\quad \textbf{MA-SAE (ours)}
& \textbf{5.9} & \textbf{71.1} & \textbf{45.4} & \textbf{3.9}
& \textbf{19.33} & \textbf{5.62}
& \textbf{27.40} & \textbf{9.59} \\

\bottomrule
\end{tabular}

\caption{
Generalization to on-policy preference tuning. We generate candidate responses from the current policy. MA-SAE consistently achieves the best performance across metrics and backbones, demonstrating that its gains extend to the on-policy setting.
}
\label{tab:hallu-onpolicy}
\end{table}

\textbf{Ablation of Modality-Specific Banks.}
Finally, we examine whether separating modality-specific features is needed for the grounding signal. We compare the full MA-SAE against a variant without the vision- and text-specific banks (full results in Appendix~\ref{app:preference-ablation}). Removing the modality-specific banks consistently degrades hallucination metrics across all three backbones, indicating that explicitly isolating modality-specific information is important for a reliable cross-modal grounding signal.

\section{Conclusion, Limitations and Future Work}
\label{sec:discussion}

We introduce MA-SAE, a modality-aware SAE that enables grounding-based preference labeling to mitigate hallucinations in MLLMs. Motivated by the observation that LLM-judge preference labels can favor linguistic style over visual grounding, MA-SAE explicitly separates vision-specific, text-specific, and cross-modal features, allowing such text-specific patterns to be isolated while using the cross-modal bank to directly measure image--text grounding. Across three MLLM models, MA-SAE consistently learns more coherent cross-modal representations, and its preference labels substantially reduce hallucinations under DPO. Its benefits further extend to mDPO and on-policy preference~tuning, suggesting that the learned grounding signal generalizes across preference-optimization objectives.

This study also has several limitations. First, our current grounding formulation and evaluation primarily focus on object-level hallucinations. Multimodal hallucinations can also involve errors in attributes, relations, and knowledge-dependent claims \citep{wang2023amber, guan2024hallusionbench, zhang2026chromouvqa}. \revision{Although our current formulation focuses on object-level grounding, MA-SAE features are extracted from the MLLM's contextualized hidden states rather than individual words. Therefore, two sentences mentioning the same objects but conveying different semantics (e.g., ``a man feeding a horse'' vs.\ ``a horse feeding a man'') may produce different feature activations and potentially different grounding scores. Evaluating and extending MA-SAE to address these broader forms of hallucination remains an important direction for future work.} Second, our experiments consider models up to 13B parameters, and it remains an open question whether the same feature decomposition and grounding signal remain effective in substantially larger models. Lastly, our current formulation focuses on vision and text. Future work could investigate whether the same modality-specific and shared feature decomposition extends to broader multimodal settings, including audio, video, tactile, metadata, and multi-sensor data \citep{rubenstein2023audiopalm, ren2024timechat, yue2025bimodal, cai2025multimodal, xu2026mobifusiondrm, xu2026rivatfuse, wu2026generative, Meng_2026_CVPR, caring2026ozchi}.

\bibliography{references}
\bibliographystyle{iclr2027_conference}


\clearpage
\appendix
\onecolumn
\renewcommand\thepart{}
\renewcommand\partname{}
\part{Appendix}
\setcounter{secnumdepth}{3}
\setcounter{tocdepth}{3}
{
\hypersetup{
    linkcolor=blue,
    pdfborder={0 0 0}
}
\parttoc
}
\newpage
\section{Linguistic Bias in LLM-Derived Preferences}
\label{app:linguistic-bias}

\subsection{Style Injection for the LLM-Judge Probe}
\label{app:style-injection}

To test whether LLM judges are sensitive to linguistic style independently of visually grounded content, we construct a controlled style-injection intervention on the widely used
MMHal-Bench \citep{sun2024aligning}. We choose MMHal-Bench because its evaluation relies on an LLM judge to score model responses, making it a natural testbed for probing whether stylistic changes can influence multimodal hallucination evaluation.

We use the 96 MMHal-Bench questions and the corresponding responses generated by the untuned LLaVA-1.5-7B baseline. We apply a deterministic rule-based style injection designed to modify only surface-level linguistic style.

\paragraph{The Style Injection.}
We apply three families of edits, none of which introduces a new verifiable claim about the image.

{\sloppy
\begin{enumerate}[leftmargin=1.5em]
\itemsep2pt

\item \emph{Hedging.}
Eight regular-expression rules convert assertions into hedged versions of the
same assertions. For example, ``there is X'' becomes ``there appears to be X'';
``the \textit{noun} is'' becomes ``the \textit{noun} appears to be''; and
``I can see'' becomes ``I can observe what appears to be.''

\item \emph{Preamble.}
We prepend the fixed opener ``Based on the visual information in the image,''
unless the response already begins with a similar phrase. This addition
restates that the response is describing the image without adding new
image-specific content.

\item \emph{Stylistic filler.}
We append one sentence from a closed set of five stylistic filler sentences
that do not introduce verifiable image content, such as ``The image captures
this moment with clarity, allowing for a detailed observation.''

\end{enumerate}
\par}

Importantly, the style injection changes only linguistic presentation by making
the response more verbose and hedged; it does not add or remove objects,
attributes, or other image-grounded claims. The example below illustrates the response before and after
style injection.

\begin{table}[h]
\centering
\small
\begin{tabular}{@{}lp{0.82\linewidth}@{}}
\toprule
\textsc{Original}
& No one is sitting on the bench. It is empty. \\

\addlinespace

\textsc{Style-injected}
& \textbf{Based on the visual information in the image,} no one
\textbf{appears to be} sitting on the bench. It \textbf{appears to be} empty.
\textbf{The image captures this moment with clarity, allowing for a detailed
observation.} \\

\bottomrule
\end{tabular}
\end{table}

\paragraph{LLM-Judge Scoring.}
We evaluate the original and style-injected responses using three LLM~judges: Gemini-3.8-Flash, Claude Opus 5, and GPT-5.5. For each judge, we use the MMHal-Bench evaluation prompts unchanged and score the same 96 response pairs under the original and style-injected conditions. Table~\ref{tab:judge-style} shows that all three judges assign higher scores to the style-injected responses
and flag them as hallucinated less often, despite the underlying factual content being unchanged. This provides direct evidence that LLM-judge evaluations are sensitive to linguistic style, favoring stylistically preferred responses even when the underlying image-grounded content is held fixed.

\clearpage
\subsection{Preference Labels Favor Hedged Responses}
\label{app:preference-labels-favor}

Building on the finding that LLM judges can reward more verbose and hedged responses even when visually grounded content is held fixed (Table~\ref{tab:judge-style}), we next examine whether this stylistic preference is reflected in preference pairs labeled with LLM-judge ratings.

We analyze VLFeedback \citep{li2024vlfeedback}, a widely used multimodal preference dataset whose response ratings are provided by GPT-4V. To mitigate differences arising from response length, we restrict the analysis to preference pairs whose preferred and rejected responses differ in word count by at most $15\%$. As shown in Table~\ref{tab:preference-hedging}, preferred responses contain substantially more hedging than rejected responses across nearly all expressions, with $1.48\times$ more hedge occurrences overall. This suggests that the stylistic preferences of the LLM judge are reflected in the resulting preference labels when its ratings are used to construct the dataset.

\begin{table}[h]
\centering
\small
\setlength{\tabcolsep}{6pt}
\renewcommand{\arraystretch}{1.08}

\begin{tabular}{l rrr}
\toprule
& Preferred & Rejected & Ratio \\
\midrule

\multicolumn{4}{l}{\textit{Phrase-level occurrences}} \\[1pt]

``suggest(s) / suggesting''
& 549 & 321 & $1.71\times$ \\

``appear(s) to (be)''
& 381 & 206 & $1.85\times$ \\

``could be''
& 231 & 193 & $1.20\times$ \\

``might be''
& 252 & 220 & $1.15\times$ \\

``indicate(s) / indicating''
& 367 & 256 & $1.43\times$ \\

``likely''
& 260 & 196 & $1.33\times$ \\

``possibly''
& 241 & 165 & $1.46\times$ \\

``may be''
& 85 & 50 & $1.70\times$ \\

other hedges
& 298 & 187 & $1.59\times$ \\

\midrule
\multicolumn{4}{l}{\textit{Summary}} \\[1pt]

All hedges
& \textbf{2664} & \textbf{1794} & $\mathbf{1.48\times}$ \\

Responses with hedge (\%)
& 60.6 & 55.5 & $1.09\times$ \\

\bottomrule
\end{tabular}

\caption{
LLM-derived preference labels favor hedged wording. After controlling for response length, preferred responses still contain substantially more hedging than rejected responses, suggesting that LLM-derived preference signals favor linguistic style beyond length alone.
}
\label{tab:preference-hedging}
\end{table}

\clearpage
\subsection{DPO Propagates Hedging into Model Outputs}
\label{app:dpo-propagates}
We next analyze whether DPO training on LLM-derived preference data transfers linguistic biases from the judge model into the optimized model, focusing on hedging and uncertainty expressions. We compare LLaVA-1.5-7B before and after DPO across AMBER, Object HalBench, and MMHal-Bench, measuring both phrase-level hedge occurrences and the fraction of responses containing at least one hedge. As shown in Table~\ref{tab:hedging-phrases}, DPO consistently increases hedging across all three benchmarks, suggesting that stylistic preferences encoded by the LLM judge can be propagated through preference optimization into the trained model.

\begin{table}[h]
\centering
\small
\setlength{\tabcolsep}{4.5pt}
\renewcommand{\arraystretch}{1.08}

\begin{tabular}{l rr rr rr r}
\toprule
& \multicolumn{2}{c}{AMBER}
& \multicolumn{2}{c}{Obj. HalBench}
& \multicolumn{2}{c}{MMHal-Bench}
& \multicolumn{1}{c}{Ratio} \\
\cmidrule(lr){2-3}
\cmidrule(lr){4-5}
\cmidrule(lr){6-7}
& Base & DPO & Base & DPO & Base & DPO & \\
\midrule

\multicolumn{8}{l}{\textit{Phrase-level occurrences}} \\[1pt]

appear(s) to (be)
& 649 & 1252
& 143 & 376
& 6 & 53
& $2.1\times$ \\

possibly
& 348 & 414
& 128 & 168
& 5 & 11
& $1.2\times$ \\

seem(s) to (be)
& 82 & 315
& 17 & 83
& 1 & 9
& $4.1\times$ \\

suggest(s) / suggesting
& 78 & 233
& 37 & 112
& 3 & 19
& $3.1\times$ \\

likely
& 89 & 232
& 49 & 113
& 2 & 17
& $2.6\times$ \\

indicate(s) / indicating
& 37 & 174
& 28 & 69
& 2 & 8
& $3.7\times$ \\

slightly
& 62 & 182
& 10 & 42
& 0 & 5
& $3.2\times$ \\

relatively
& 20 & 76
& 1 & 12
& 0 & 3
& $4.3\times$ \\

other hedges
& 48 & 197
& 28 & 71
& 1 & 17
& $3.7\times$ \\

\midrule

\multicolumn{8}{l}{\textit{Summary}} \\[1pt]

All hedges
& 1413 & \textbf{3075}
& 441  & \textbf{1046}
& 20   & \textbf{142}
& $2.3\times$ \\

Responses with hedge (\%)
& 79 & \textbf{98}
& 77 & \textbf{100}
& 11 & \textbf{55}
& $1.29\times$ \\

\bottomrule
\end{tabular}

\caption{
DPO on LLM-derived preferences substantially increases hedging. We compare LLaVA-1.5-7B before and after DPO training on preference pairs labeled using LLM-judge ratings. Hedging increases consistently
across all three hallucination benchmarks, both in phrase-level occurrences
and in the fraction of responses containing hedging.
}
\label{tab:hedging-phrases}
\end{table}

\clearpage
\subsection{MA-SAE Preference Labeling Does Not Favor Hedging}
\label{app:masae-preference-hedging}

We next repeat the analysis of Appendix~\ref{app:preference-labels-favor} using preference pairs labeled by the MA-SAE grounding score rather than LLM-judge ratings. As shown in Table~\ref{tab:masae-preference-hedging}, preferred and rejected responses exhibit nearly identical levels of hedging overall ($1.02\times$), with no consistent preference across individual expressions. In contrast to LLM-derived labels, MA-SAE preference labeling does not systematically favor hedged wording.

\begin{table}[h]
\centering
\small
\setlength{\tabcolsep}{6pt}
\renewcommand{\arraystretch}{1.08}

\begin{tabular}{l rrr}
\toprule
& Preferred & Rejected & Ratio \\
\midrule

\multicolumn{4}{l}{\textit{Phrase-level occurrences}} \\[1pt]

``suggest(s) / suggesting''
& 601 & 503 & $1.19\times$ \\

``appear(s) to (be)''
& 329 & 368 & $0.89\times$ \\

``could be''
& 409 & 398 & $1.03\times$ \\

``might be''
& 410 & 408 & $1.00\times$ \\

``indicate(s) / indicating''
& 368 & 357 & $1.03\times$ \\

``likely''
& 337 & 307 & $1.10\times$ \\

``possibly''
& 287 & 299 & $0.96\times$ \\

``may be''
& 110 & 116 & $0.95\times$ \\

other hedges
& 181 & 221 & $0.82\times$ \\

\midrule
\multicolumn{4}{l}{\textit{Summary}} \\[1pt]

All hedges
& \textbf{3032} & \textbf{2977} & $\mathbf{1.02\times}$ \\

Responses with hedge (\%)
& 45.9 & 47.1 & $0.98\times$ \\

\bottomrule
\end{tabular}

\caption{
MA-SAE preference labeling does not favor hedged wording. After controlling for response length, preferred and rejected responses exhibit nearly identical levels of hedging, in contrast to preferences labeled with GPT-4V ratings.
}
\label{tab:masae-preference-hedging}
\end{table}

\clearpage
\section{MA-SAE Architecture and Training Details}
\label{appendix:masae-training}
Table~\ref{tab:masae-hparams} summarizes the MA-SAE architecture and training hyperparameters for the three backbones. For a controlled comparison, all SAE baselines (Vanilla SAE, VL-SAE, and GSAE) are trained on the same backbone layer and LLaVA-Instruct-150K training split, and evaluated on the same held-out split. Each baseline uses the same dictionary size $p$ and per-token sparsity budget $K=64$ as MA-SAE, ensuring matched parameter and sparsity budgets across methods.

\vspace{0.3cm}
\begin{table}[h]
\centering
\small
\setlength{\tabcolsep}{5pt}
\renewcommand{\arraystretch}{1.08}
\begin{tabular}{lccc}
\toprule
\textbf{Hyperparameter}
& \textbf{LLaVA-1.5-7B}
& \textbf{LLaVA-1.5-13B}
& \textbf{InternVL3-8B} \\
\midrule
\multicolumn{4}{c}{\textbf{MA-SAE Architecture}} \\
\midrule
SAE layer
& 16
& 20
& 14 \\
Input dimension
& 4096
& 5120
& 3584 \\
Expansion factor
& $5\times$
& $5\times$
& $5\times$ \\
Dictionary size $p$
& 20{,}480
& 25{,}600
& 17{,}920 \\
\quad Vision-only bank
& 6{,}144 (30\%)
& 7{,}680 (30\%)
& 5{,}376 (30\%) \\
\quad Text-only bank
& 6{,}144 (30\%)
& 7{,}680 (30\%)
& 5{,}376 (30\%) \\
\quad Cross-modal bank
& 8{,}192 (40\%)
& 10{,}240 (40\%)
& 7{,}168 (40\%) \\
\midrule
\multicolumn{4}{c}{\textbf{Shared Training Settings}} \\
\midrule
Activation
& \multicolumn{3}{c}{ReLU + Top-$K$} \\
Top-$K$ budget
& \multicolumn{3}{c}{$K=64$} \\
\quad $K_{\mathrm{mod}}$
& \multicolumn{3}{c}{27} \\
\quad $K_{\mathrm{cross}}$
& \multicolumn{3}{c}{37} \\
Learning rate
& \multicolumn{3}{c}{$5\times10^{-5}$} \\
Warmup steps
& \multicolumn{3}{c}{1{,}000} \\
LR schedule
& \multicolumn{3}{c}{Warmup $\rightarrow$ constant} \\
Training epochs
& \multicolumn{3}{c}{3} \\
Batch size
& \multicolumn{3}{c}{32 samples/step} \\
Adam $\beta_1$, $\beta_2$
& \multicolumn{3}{c}{0.9, 0.999} \\
Total training steps
& \multicolumn{3}{c}{32{,}967} \\
$\lambda_{\mathrm{BT}}$
& \multicolumn{3}{c}{5.0} \\
Barlow Twins off-diagonal $\lambda_{\mathrm{off}}$
& \multicolumn{3}{c}{0.005} \\
Alignment warmup steps
& \multicolumn{3}{c}{5{,}000} \\
\bottomrule
\end{tabular}
\caption{MA-SAE architecture and training hyperparameters.}
\label{tab:masae-hparams}
\end{table}

\clearpage
\section{Qualitative Analysis of Learned MA-SAE Features}\label{appendix: examples}
\raggedbottom
Qualitative analysis of the features learned by MA-SAE at layer~16 of
LLaVA-1.5-7B, trained on LLaVA-Instruct-150K activations and evaluated on a held-out split. We show features from all three banks: vision-specific, text-specific, and the shared cross-modal bank. For each feature, we present its top-activating examples,
visualized as heatmaps over the image patch grid for vision activations and per-token shading for text activations. We observe qualitatively consistent feature patterns in MA-SAE dictionaries trained on LLaVA-1.5-13B and InternVL3-8B.



\subsection{Cross-modal Bank}

\textbf{Cross-modal bank: features encode shared concepts across modalities.}

Each block shows one feature from the cross-modal bank. The images show its strongest vision-side activations as heatmaps over the patch grid, while the corresponding responses show its strongest text-side activations with per-token shading.

\medskip

\addvspace{4pt}

\noindent\begin{minipage}{\linewidth}
\qfeat{bankbind}{17870}{double-decker bus}
\noindent \qimg{0.2350\linewidth}{}{1}\hfill \qimg{0.2350\linewidth}{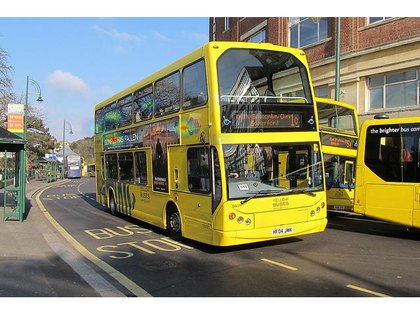}{2}\hfill \qimg{0.2350\linewidth}{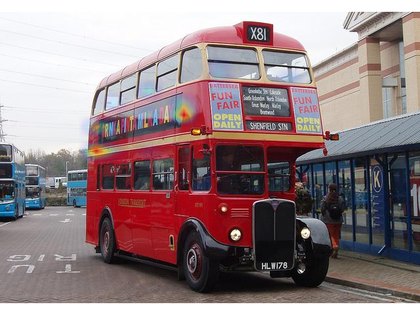}{3}\hfill \qimg{0.2350\linewidth}{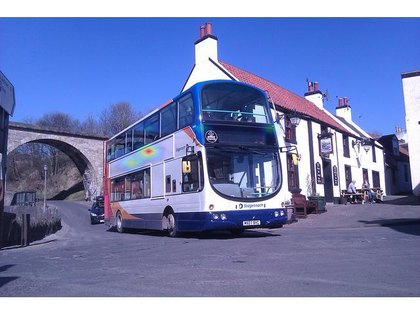}{4}\par\nobreak
\end{minipage}\par\nobreak
\vspace{3pt}\nobreak
\noindent{\footnotesize\hangindent=1.4em\hangafter=1 \qnum{1}\enspace  \qhl{56}{Double-decker} buses, with their distinctive two-level design, have been an iconic symbol of        \qelide{}\par}
\nobreak\vspace{1pt}
\noindent{\footnotesize\hangindent=1.4em\hangafter=1 \qnum{2}\enspace  A \qhl{40}{double-decker} bus is a type of bus that has two levels or decks,             \qelide{}\par}
\penalty200\vspace{1pt}
\noindent{\footnotesize\hangindent=1.4em\hangafter=1 \qnum{3}\enspace  A double \qhl{40}{decker} bus offers a unique and enhanced experience for passengers compared to regular      \qelide{}\par}
\penalty200\vspace{1pt}
\noindent{\footnotesize\hangindent=1.4em\hangafter=1 \qnum{4}\enspace \qelide{} as a contrast to the busy city life with vehicles like the \qhl{61}{double-decker} bus and cars driving around, creating a balance between urban and natural\qelide{}\par}
\penalty200\vspace{1pt}
\par\addvspace{7pt}

\noindent\begin{minipage}{\linewidth}
\qfeat{bankbind}{18556}{zebra}
\noindent \qimg{0.2350\linewidth}{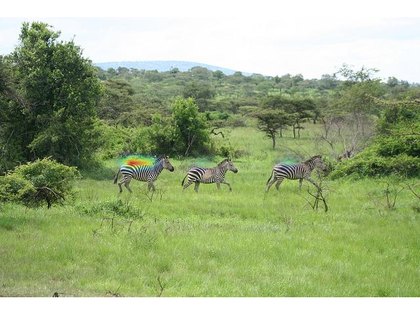}{1}\hfill \qimg{0.2350\linewidth}{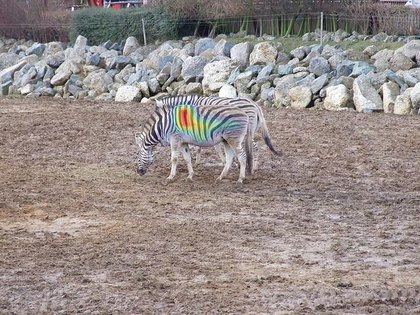}{2}\hfill \qimg{0.2350\linewidth}{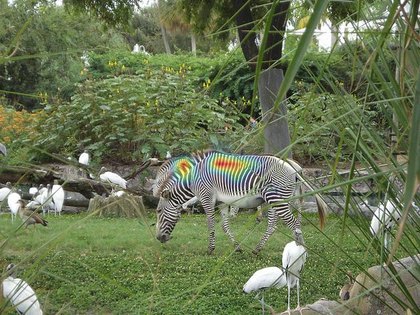}{3}\hfill \qimg{0.2350\linewidth}{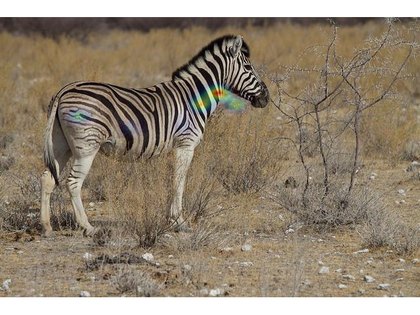}{4}\par\nobreak
\end{minipage}\par\nobreak
\vspace{3pt}\nobreak
\noindent{\footnotesize\hangindent=1.4em\hangafter=1 \qnum{1}\enspace  Preserving the \qhl{70}{zebra's} natural habitat involves various factors that help maintain a healthy ecosystem in                 \qelide{}\par}
\nobreak\vspace{1pt}
\noindent{\footnotesize\hangindent=1.4em\hangafter=1 \qnum{2}\enspace \qelide{} in a zoo, wildlife park, or sanctuary. In such facilities, animals like \qhl{62}{zebras} are kept in contained areas to ensure their safety, the safety of    \qelide{}\par}
\penalty200\vspace{1pt}
\noindent{\footnotesize\hangindent=1.4em\hangafter=1 \qnum{3}\enspace  In their natural habitat, \qhl{59}{zebras} typically graze on grasses and other vegetation. Their diet primarily consists of    \qelide{}\par}
\penalty200\vspace{1pt}
\noindent{\footnotesize\hangindent=1.4em\hangafter=1 \qnum{4}\enspace \qelide{} welfare advocates might have concerns regarding the living conditions of the two \qhl{34}{zebras} seen in the image. Their enclosure, which consists of paved surfaces and  \qelide{}\par}
\penalty200\vspace{1pt}
\par\addvspace{7pt}

\noindent\begin{minipage}{\linewidth}
\qfeat{bankbind}{15849}{giraffe}
\noindent \qimg{0.2350\linewidth}{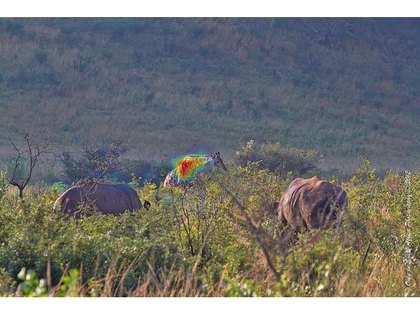}{1}\hfill \qimg{0.2350\linewidth}{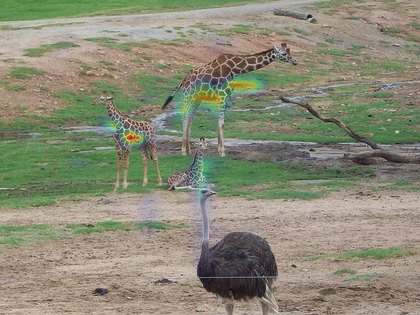}{2}\hfill \qimg{0.2350\linewidth}{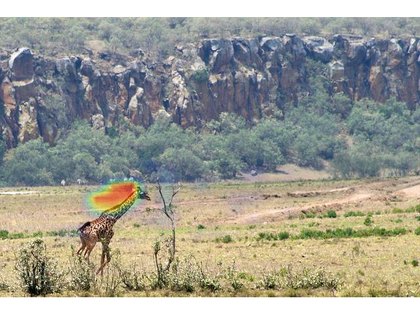}{3}\hfill \qimg{0.2350\linewidth}{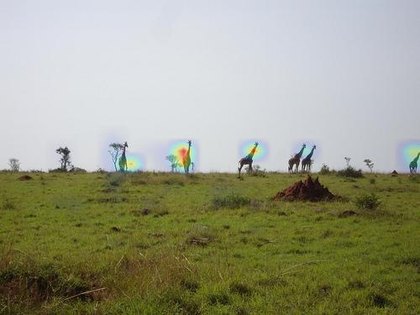}{4}\par\nobreak
\end{minipage}\par\nobreak
\vspace{3pt}\nobreak
\noindent{\footnotesize\hangindent=1.4em\hangafter=1 \qnum{1}\enspace \qelide{} a grassland or field environment, which is a natural habitat for both \qhl{70}{giraffes} and zebras.\par}
\nobreak\vspace{1pt}
\noindent{\footnotesize\hangindent=1.4em\hangafter=1 \qnum{2}\enspace \qelide{} picture was taken at a wildlife park, zoo, or sanctuary, as the \qhl{69}{giraffes} are enclosed in a fenced area, hinting at a man-made environment designed\qelide{}\par}
\penalty200\vspace{1pt}
\noindent{\footnotesize\hangindent=1.4em\hangafter=1 \qnum{3}\enspace  An ostrich might be housed in the same enclosure as \qhl{56}{giraffes} in a zoo because both animals are native to similar habitats, primarily\qelide{}\par}
\penalty200\vspace{1pt}
\noindent{\footnotesize\hangindent=1.4em\hangafter=1 \qnum{4}\enspace \qelide{} or savanna, which is a characteristic environment for \qhl{53}{giraffes} in the wild. \qhl{64}{Giraffe} populations are predominantly found in sub-Saharan African countries, spanning across countries like    \qelide{}\par}
\penalty200\vspace{1pt}
\par\addvspace{7pt}

\noindent\begin{minipage}{\linewidth}
\qfeat{bankbind}{17382}{elephant}
\noindent \qimg{0.2350\linewidth}{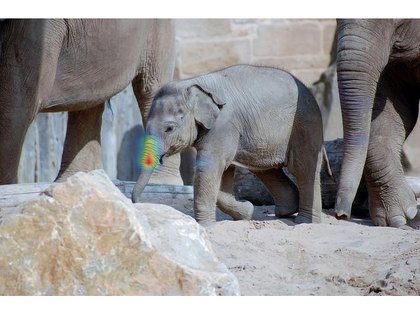}{1}\hfill \qimg{0.2350\linewidth}{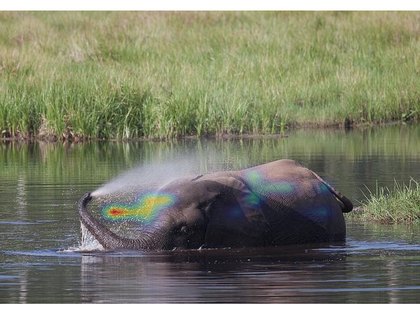}{2}\hfill \qimg{0.2350\linewidth}{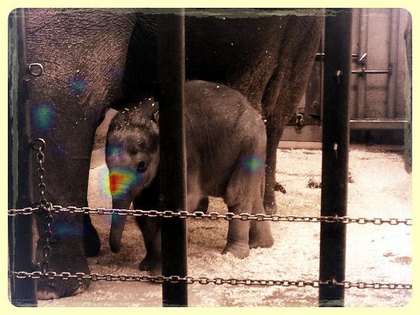}{3}\hfill \qimg{0.2350\linewidth}{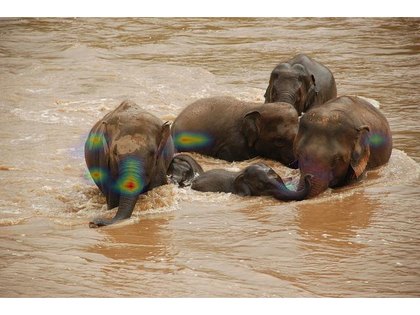}{4}\par\nobreak
\end{minipage}\par\nobreak
\vspace{3pt}\nobreak
\noindent{\footnotesize\hangindent=1.4em\hangafter=1 \qnum{1}\enspace  In the image, there are two \qhl{35}{elephants} - a mother \qhl{33}{elephant} and a baby \qhl{24}{elephant} - standing together on\qelide{}\par}
\nobreak\vspace{1pt}
\noindent{\footnotesize\hangindent=1.4em\hangafter=1 \qnum{2}\enspace  \qhl{44}{Elephant} habitats and conservation efforts are crucial to the survival and well-being of                          \qelide{}\par}
\penalty200\vspace{1pt}
\noindent{\footnotesize\hangindent=1.4em\hangafter=1 \qnum{3}\enspace \qelide{} \qhl{42}{elephant} might be walking with its baby for several reasons. First, adult \qhl{56}{elephants} play a critical role in the upbringing and protection of their young,\qelide{}\par}
\penalty200\vspace{1pt}
\noindent{\footnotesize\hangindent=1.4em\hangafter=1 \qnum{4}\enspace \qelide{} engaged in shared activities such as swimming or playing in a river. \qhl{58}{Elephants} are known for their intelligence, intricate communication, and strong family ties. They    \qelide{}\par}
\penalty200\vspace{1pt}
\par\addvspace{7pt}

\noindent\begin{minipage}{\linewidth}
\qfeat{bankbind}{15772}{sheep}
\noindent \qimg{0.2350\linewidth}{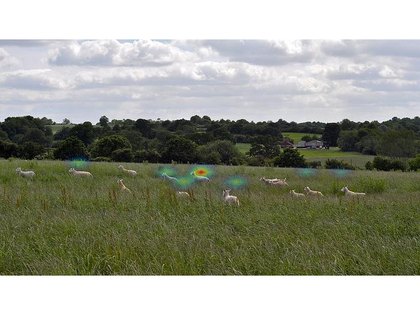}{1}\hfill \qimg{0.2350\linewidth}{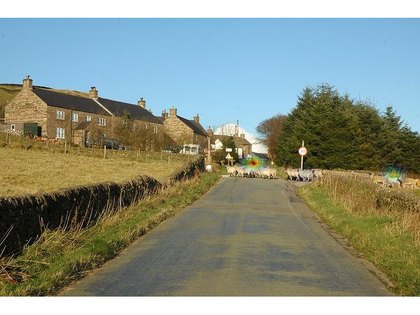}{2}\hfill \qimg{0.2350\linewidth}{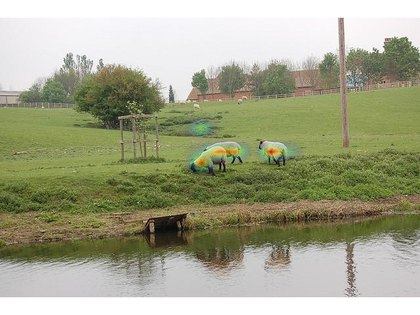}{3}\hfill \qimg{0.2350\linewidth}{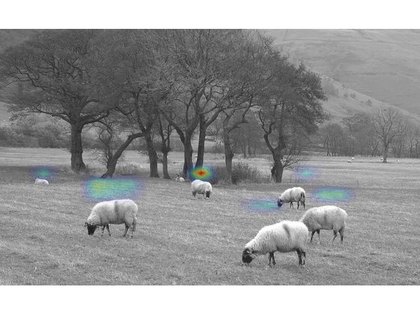}{4}\par\nobreak
\end{minipage}\par\nobreak
\vspace{3pt}\nobreak
\noindent{\footnotesize\hangindent=1.4em\hangafter=1 \qnum{1}\enspace  The \qhl{57}{sheep} in the foggy and rocky environment might face several challenges:                 \qelide{}\par}
\nobreak\vspace{1pt}
\noindent{\footnotesize\hangindent=1.4em\hangafter=1 \qnum{2}\enspace  There are several reasons for raising \qhl{52}{sheep} in a field with tall grass:    Natural diet: \qhl{64}{Sheep}           \qelide{}\par}
\penalty200\vspace{1pt}
\noindent{\footnotesize\hangindent=1.4em\hangafter=1 \qnum{3}\enspace       \qelide{} maintains a balanced ecosystem.    Soil fertility and erosion control: \qhl{53}{Sheep's} natural grazing behavior contributes to soil health, as the even distribution of  \qelide{}\par}
\penalty200\vspace{1pt}
\noindent{\footnotesize\hangindent=1.4em\hangafter=1 \qnum{4}\enspace  The presence of the \qhl{64}{sheep} grazing in the grass-covered field suggests a rural, agricultural, or pastoral setting.}
\penalty200\vspace{1pt}
\par\addvspace{7pt}

\noindent\begin{minipage}{\linewidth}
\qfeat{bankbind}{13809}{pizza}
\noindent \qimg{0.2350\linewidth}{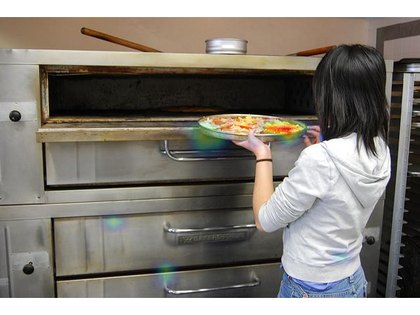}{1}\hfill \qimg{0.2350\linewidth}{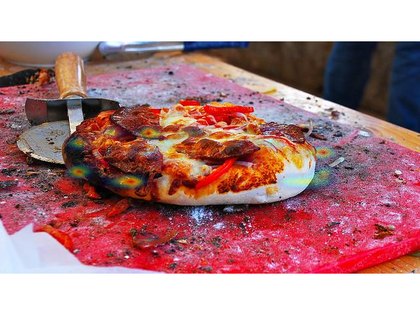}{2}\hfill \qimg{0.2350\linewidth}{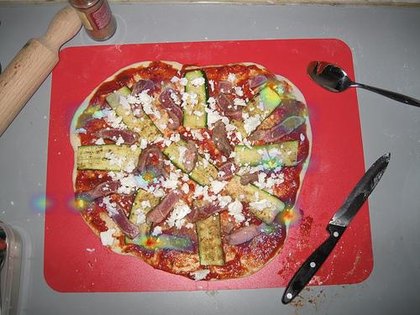}{3}\hfill \qimg{0.2350\linewidth}{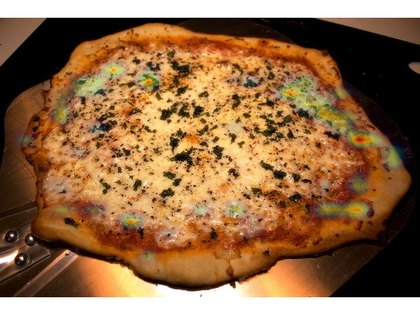}{4}\par\nobreak
\end{minipage}\par\nobreak
\vspace{3pt}\nobreak
\noindent{\footnotesize\hangindent=1.4em\hangafter=1 \qnum{1}\enspace  The \qhl{70}{pizza} with cheese and black olives in the image could be suitable for  \qelide{}\par}
\nobreak\vspace{1pt}
\noindent{\footnotesize\hangindent=1.4em\hangafter=1 \qnum{2}\enspace  Several factors could have contributed to the burnt \qhl{25}{crust} of the \qhl{67}{pizza} in the image. These factors may include:    Overcooking: The        \qelide{}\par}
\penalty200\vspace{1pt}
\noindent{\footnotesize\hangindent=1.4em\hangafter=1 \qnum{3}\enspace  To enhance the flavor of the cheese \qhl{66}{pizza,} one can add various toppings and seasonings to suit their preferences.\par}
\penalty200\vspace{1pt}
\noindent{\footnotesize\hangindent=1.4em\hangafter=1 \qnum{4}\enspace  The appeal of the \qhl{58}{pizza} in the image primarily comes from its presentation, variety of toppings, and\qelide{}\par}
\penalty200\vspace{1pt}
\par\addvspace{7pt}

\noindent\begin{minipage}{\linewidth}
\qfeat{bankbind}{18036}{cake}
\noindent \qimg{0.2350\linewidth}{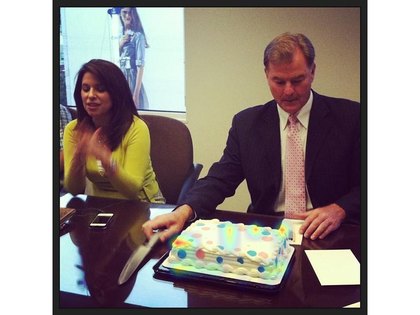}{1}\hfill \qimg{0.2350\linewidth}{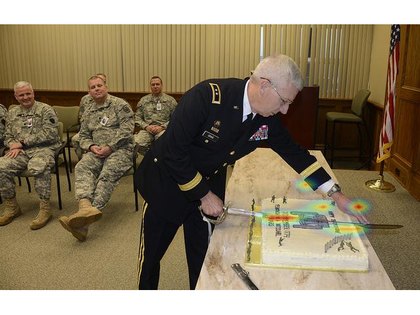}{2}\hfill \qimg{0.2350\linewidth}{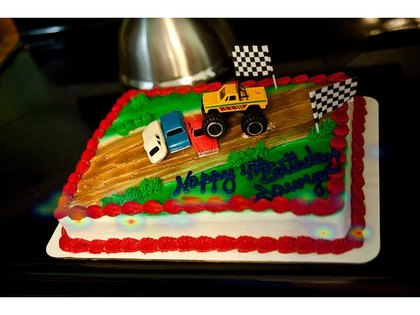}{3}\hfill \qimg{0.2350\linewidth}{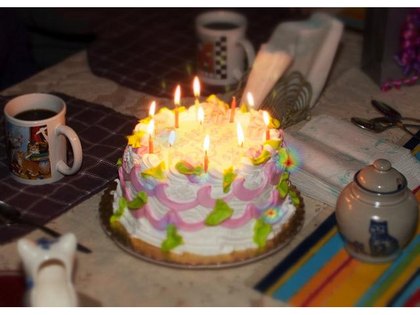}{4}\par\nobreak
\end{minipage}\par\nobreak
\vspace{3pt}\nobreak
\noindent{\footnotesize\hangindent=1.4em\hangafter=1 \qnum{1}\enspace \qelide{} to determine the exact occasion based on the image alone, the decorated \qhl{70}{cake} could be for a variety of occasions, such as a birthday, anniversary,\qelide{}\par}
\nobreak\vspace{1pt}
\noindent{\footnotesize\hangindent=1.4em\hangafter=1 \qnum{2}\enspace  When planning to use this beautifully designed \qhl{59}{cake} for a special occasion, several considerations should be made:                  \qelide{}\par}
\penalty200\vspace{1pt}
\noindent{\footnotesize\hangindent=1.4em\hangafter=1 \qnum{3}\enspace  The elaborate decoration of the \qhl{62}{cake,} which features a woman in a carriage and a bride and groom\qelide{}\par}
\penalty200\vspace{1pt}
\noindent{\footnotesize\hangindent=1.4em\hangafter=1 \qnum{4}\enspace  The \qhl{58}{cake} in the image is decorated with an Angry Birds theme, featuring green\qelide{}\par}
\penalty200\vspace{1pt}
\par\addvspace{7pt}

\noindent\begin{minipage}{\linewidth}
\qfeat{bankbind}{17460}{fruit}
\noindent \qimg{0.2350\linewidth}{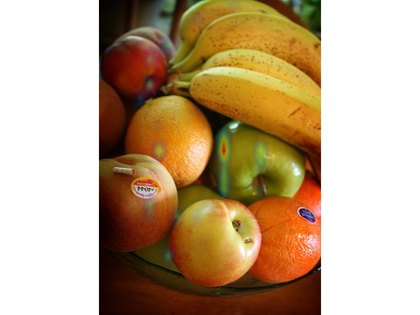}{1}\hfill \qimg{0.2350\linewidth}{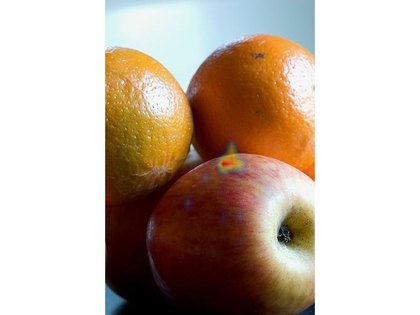}{2}\hfill \qimg{0.2350\linewidth}{}{3}\hfill \qimg{0.2350\linewidth}{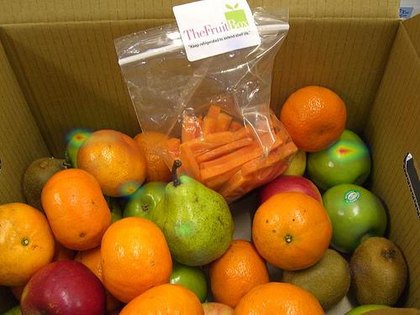}{4}\par\nobreak
\end{minipage}\par\nobreak
\vspace{3pt}\nobreak
\noindent{\footnotesize\hangindent=1.4em\hangafter=1 \qnum{1}\enspace  Consuming a diverse range of \qhl{50}{fruits} provides several health benefits, as different \qhl{40}{fruits} offer a unique mix \qelide{}\par}
\nobreak\vspace{1pt}
\noindent{\footnotesize\hangindent=1.4em\hangafter=1 \qnum{2}\enspace  The neat organization of \qhl{46}{fruits} in the market is beneficial to both sellers and customers in several  \qelide{}\par}
\penalty200\vspace{1pt}
\noindent{\footnotesize\hangindent=1.4em\hangafter=1 \qnum{3}\enspace    \qelide{} shorter distance from the farm to the point of sale. Conversely, imported \qhl{53}{fruits} and vegetables may spend more time in transit, affecting their freshness.           \qelide{}\par}
\penalty200\vspace{1pt}
\noindent{\footnotesize\hangindent=1.4em\hangafter=1 \qnum{4}\enspace  Consuming a variety of \qhl{52}{fruits} on a regular basis offers numerous health benefits, mainly due to their                 \qelide{}\par}
\penalty200\vspace{1pt}
\par\addvspace{7pt}

\noindent\begin{minipage}{\linewidth}
\qfeat{bankbind}{14445}{baseball batter}
\noindent \qimg{0.2350\linewidth}{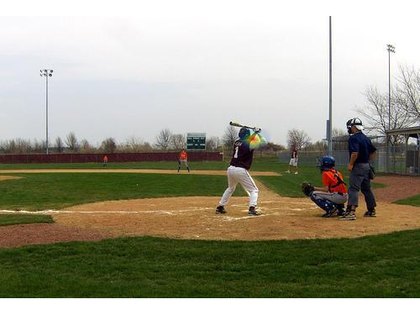}{1}\hfill \qimg{0.2350\linewidth}{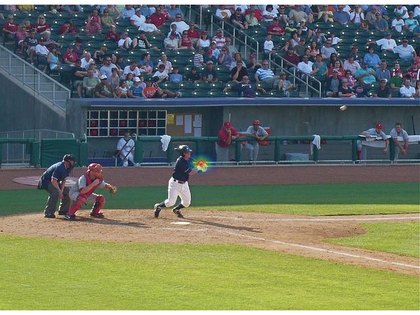}{2}\hfill \qimg{0.2350\linewidth}{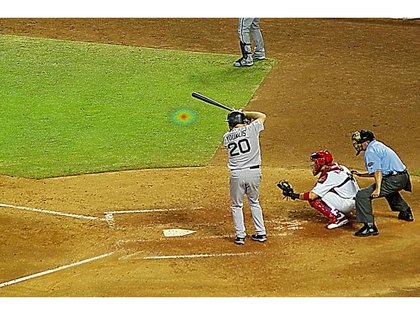}{3}\hfill \qimg{0.2350\linewidth}{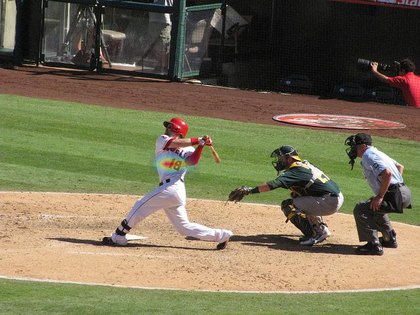}{4}\par\nobreak
\end{minipage}\par\nobreak
\vspace{3pt}\nobreak
\noindent{\footnotesize\hangindent=1.4em\hangafter=1 \qnum{1}\enspace \qelide{} situation, several factors play a crucial role in the success of the \qhl{70}{batter.} These include:    Timing: The batter's ability to anticipate the              \qelide{}\par}
\nobreak\vspace{1pt}
\noindent{\footnotesize\hangindent=1.4em\hangafter=1 \qnum{2}\enspace \qelide{} game. You can witness various elements of the game, such as the \qhl{65}{batter} at the plate with a baseball bat and a player with a\qelide{}\par}
\penalty200\vspace{1pt}
\noindent{\footnotesize\hangindent=1.4em\hangafter=1 \qnum{3}\enspace     \qelide{} out by the defense.  During their turn at bat, the \qhl{62}{batter} tries to hit the ball thrown by the pitcher. The pitcher's objective    \qelide{}\par}
\penalty200\vspace{1pt}
\noindent{\footnotesize\hangindent=1.4em\hangafter=1 \qnum{4}\enspace  When a pitch is thrown during a baseball game, the \qhl{62}{batter} has a few options for their actions: Swing \qelide{}\par}
\penalty200\vspace{1pt}
\par\addvspace{7pt}

\noindent\begin{minipage}{\linewidth}
\qfeat{bankbind}{18595}{snow sports (skiing, snowboarding)}
\noindent \qimg{0.2350\linewidth}{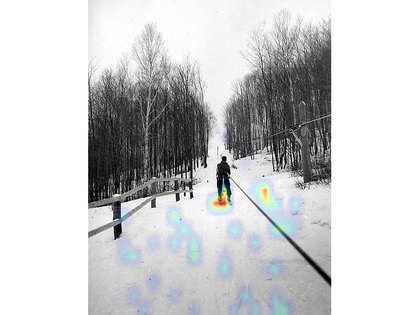}{1}\hfill \qimg{0.2350\linewidth}{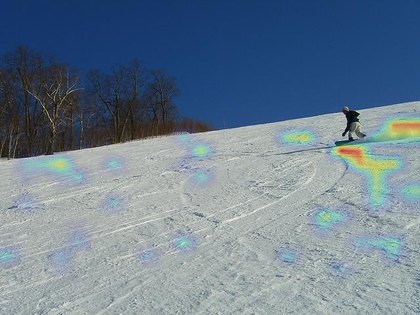}{2}\hfill \qimg{0.2350\linewidth}{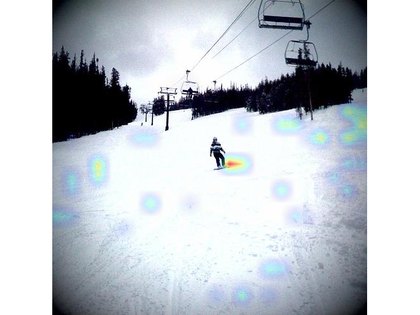}{3}\hfill \qimg{0.2350\linewidth}{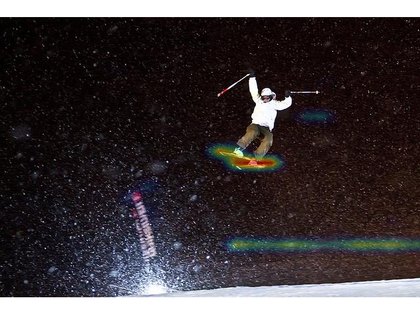}{4}\par\nobreak
\end{minipage}\par\nobreak
\vspace{3pt}\nobreak
\noindent{\footnotesize\hangindent=1.4em\hangafter=1 \qnum{1}\enspace                     \qelide{} and equipment for \qhl{47}{snowboarders} who wish to safely and enjoyably navigate the \qhl{68}{snowy} mountain terrain.\par}
\nobreak\vspace{1pt}
\noindent{\footnotesize\hangindent=1.4em\hangafter=1 \qnum{2}\enspace  While \qhl{67}{snowboarding} in a wooded area as depicted in the image, there are a                 \qelide{}\par}
\penalty200\vspace{1pt}
\noindent{\footnotesize\hangindent=1.4em\hangafter=1 \qnum{3}\enspace  \qhl{66}{Snowboarding} and performing tricks in the air can be thrilling and enjoyable, but\qelide{}\par}
\penalty200\vspace{1pt}
\noindent{\footnotesize\hangindent=1.4em\hangafter=1 \qnum{4}\enspace  When \qhl{66}{skiing} in remote areas, it is crucial for the skier to take several               \qelide{}\par}
\penalty200\vspace{1pt}
\par\addvspace{7pt}

\addvspace{15pt}
\clearpage
\subsection{Vision-specific Bank}

\textbf{Vision-specific bank: features capture modality-specific visual content.}

Each block shows one feature from the visual-only bank. The images show its strongest activations as heatmaps over the patch grid. These features are intended to capture visual content not represented in the paired text; qualitatively, they primarily respond to background, spatial, and scene-level patterns rather than salient objects mentioned in the text.

\par\addvspace{4pt}

\noindent\begin{minipage}{\linewidth}
\qfeat{bankvis}{1860}{}
\noindent \qimg{0.1840\linewidth}{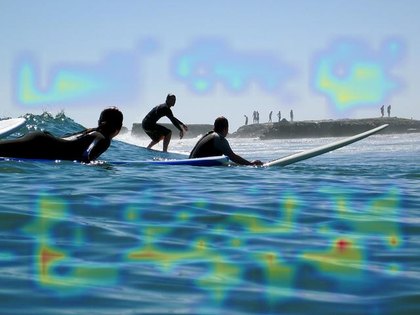}{1}\hfill \qimg{0.1840\linewidth}{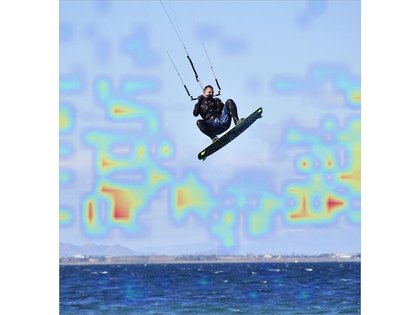}{2}\hfill \qimg{0.1840\linewidth}{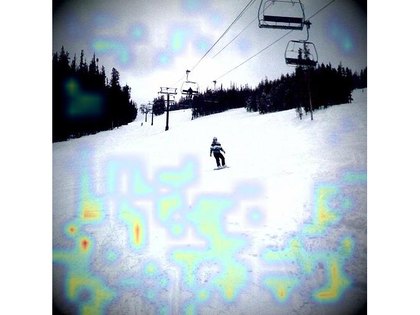}{3}\hfill \qimg{0.1840\linewidth}{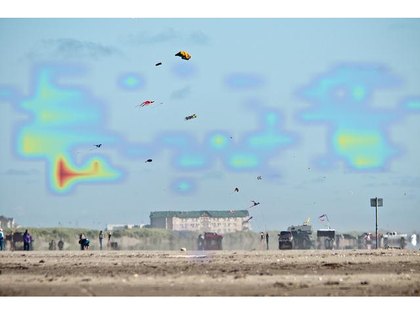}{4}\hfill \qimg{0.1840\linewidth}{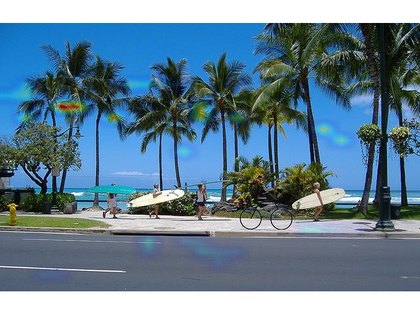}{5}\par\nobreak
\end{minipage}\par
\par\addvspace{1pt}

\noindent\begin{minipage}{\linewidth}
\qfeat{bankvis}{2020}{}
\noindent \qimg{0.1840\linewidth}{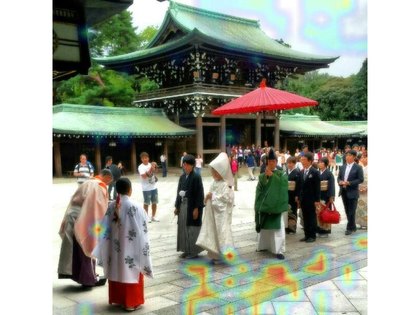}{1}\hfill \qimg{0.1840\linewidth}{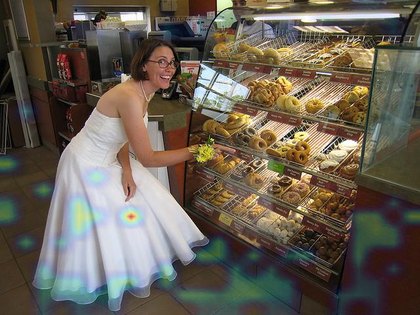}{2}\hfill \qimg{0.1840\linewidth}{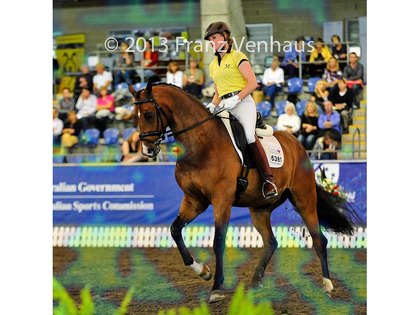}{3}\hfill \qimg{0.1840\linewidth}{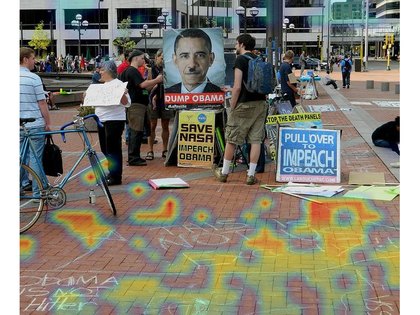}{4}\hfill \qimg{0.1840\linewidth}{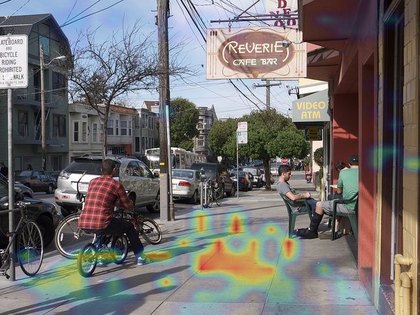}{5}\par\nobreak
\end{minipage}\par
\par\addvspace{1pt}

\noindent\begin{minipage}{\linewidth}
\qfeat{bankvis}{2411}{}
\noindent \qimg{0.1840\linewidth}{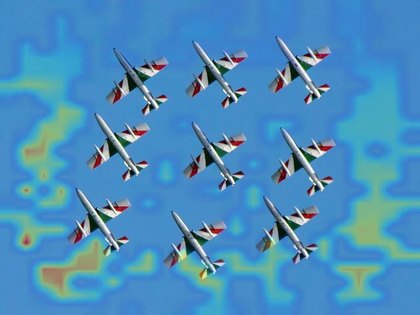}{1}\hfill \qimg{0.1840\linewidth}{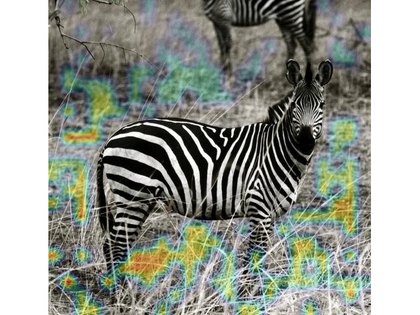}{2}\hfill \qimg{0.1840\linewidth}{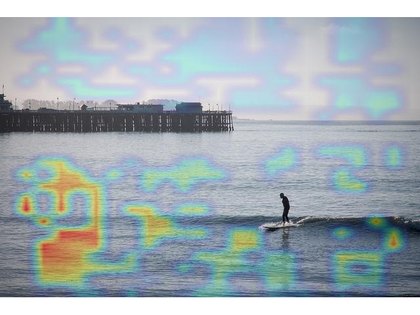}{3}\hfill \qimg{0.1840\linewidth}{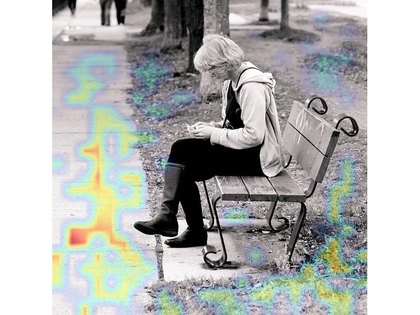}{4}\hfill \qimg{0.1840\linewidth}{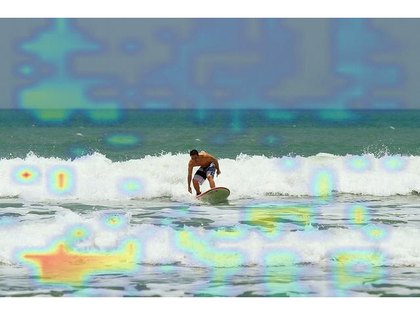}{5}\par\nobreak
\end{minipage}\par
\par\addvspace{1pt}

\noindent\begin{minipage}{\linewidth}
\qfeat{bankvis}{916}{}
\noindent \qimg{0.1840\linewidth}{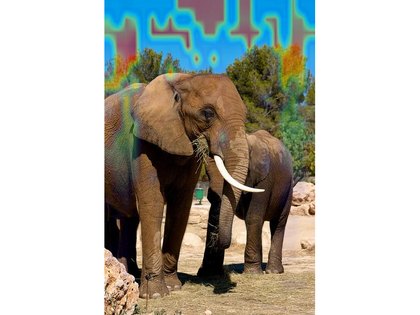}{1}\hfill \qimg{0.1840\linewidth}{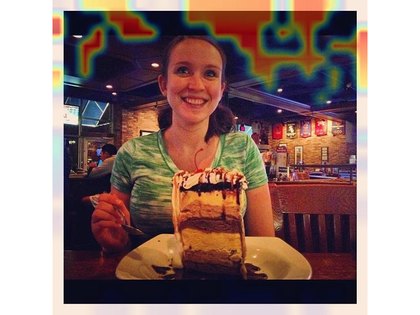}{2}\hfill \qimg{0.1840\linewidth}{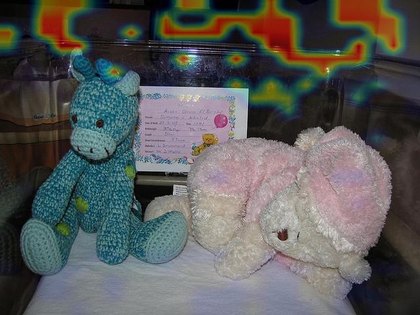}{3}\hfill \qimg{0.1840\linewidth}{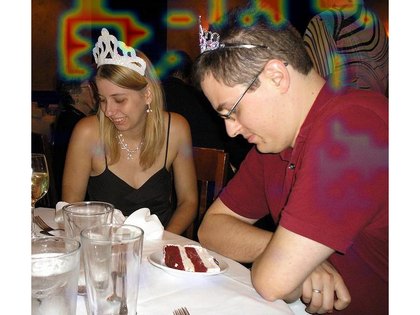}{4}\hfill \qimg{0.1840\linewidth}{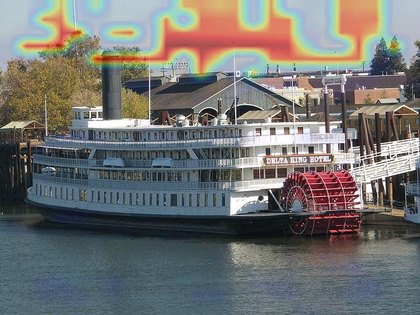}{5}\par\nobreak
\end{minipage}\par
\par\addvspace{1pt}

\noindent\begin{minipage}{\linewidth}
\qfeat{bankvis}{3208}{}
\noindent \qimg{0.1840\linewidth}{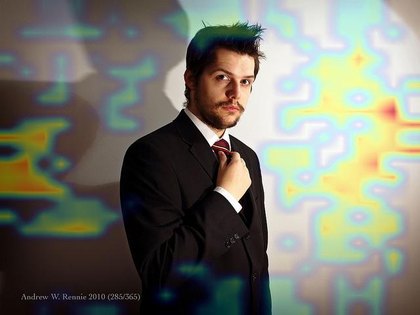}{1}\hfill \qimg{0.1840\linewidth}{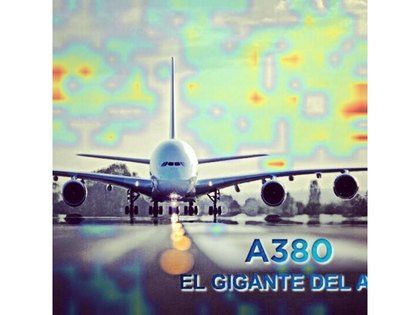}{2}\hfill \qimg{0.1840\linewidth}{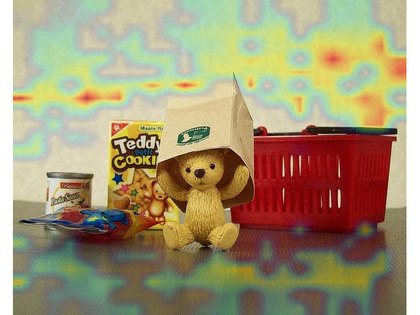}{3}\hfill \qimg{0.1840\linewidth}{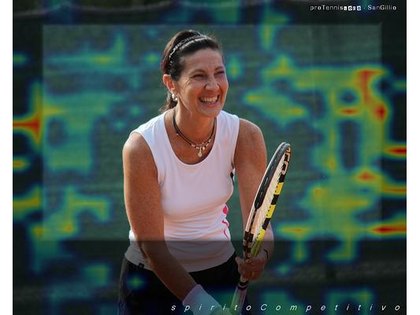}{4}\hfill \qimg{0.1840\linewidth}{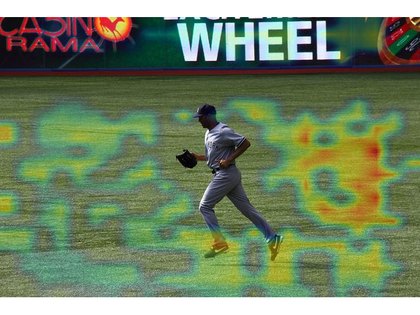}{5}\par\nobreak
\end{minipage}\par
\par\addvspace{1pt}

\noindent\begin{minipage}{\linewidth}
\qfeat{bankvis}{350}{}
\noindent \qimg{0.1840\linewidth}{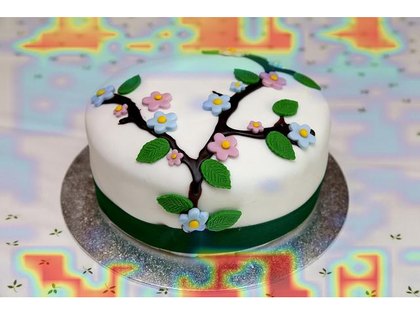}{1}\hfill \qimg{0.1840\linewidth}{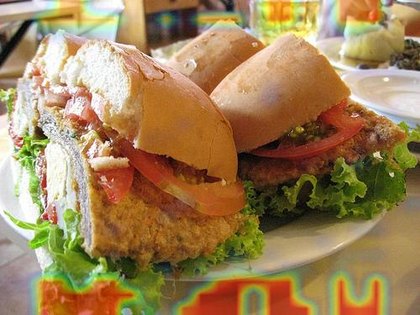}{2}\hfill \qimg{0.1840\linewidth}{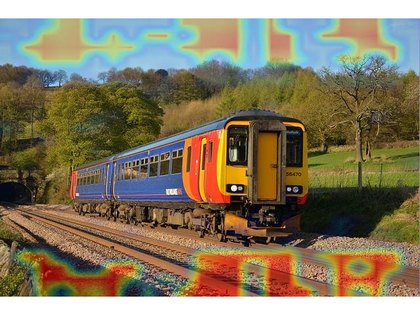}{3}\hfill \qimg{0.1840\linewidth}{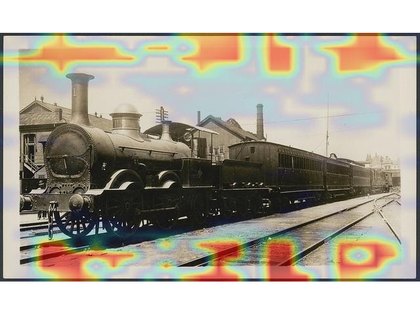}{4}\hfill \qimg{0.1840\linewidth}{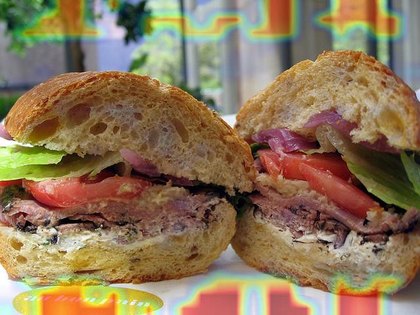}{5}\par\nobreak
\end{minipage}\par


\clearpage
\subsection{Text-specific Bank}

\textbf{Text-specific bank: features capture modality-specific linguistic patterns.}

Each block shows one feature from the text-only bank. The responses show its strongest activations with per-token shading. These features are intended to capture linguistic patterns not grounded in the paired image; qualitatively, they respond to modality-specific language such as hedging, uncertainty, and inferential expressions.

\par\addvspace{4pt}

\noindent\begin{minipage}{\linewidth}
\qfeat{banktxt}{6863}{}
\end{minipage}\par\nobreak
\vspace{3pt}\nobreak
\noindent{\footnotesize\hangindent=1.4em\hangafter=1 \qnum{1}\enspace   \qelide{} that require writing or manual note-taking. Furthermore, the cup on the table \qhl{70}{implies} that the person might be consuming a beverage, such as coffee or    \qelide{}\par}
\nobreak\vspace{1pt}
\noindent{\footnotesize\hangindent=1.4em\hangafter=1 \qnum{2}\enspace \qelide{} for tasks such as opening packages or cutting herbs. The multiple bottles \qhl{70}{imply} that they likely use various liquids, sauces, or seasonings during their cooking\qelide{}\par}
\penalty200\vspace{1pt}
\noindent{\footnotesize\hangindent=1.4em\hangafter=1 \qnum{3}\enspace \qelide{} dishes. The presence of the tray or the covered food try also \qhl{70}{implies} that the food is ready to be served or consumed, which may\qelide{}\par}
\penalty200\vspace{1pt}
\noindent{\footnotesize\hangindent=1.4em\hangafter=1 \qnum{4}\enspace \qelide{} facility that holds importance in its community. The crane in the background \qhl{70}{implies} that this area might be going through some transformation, preserving or renovating\qelide{}\par}
\penalty200\vspace{1pt}
\par\addvspace{7pt}

\noindent\begin{minipage}{\linewidth}
\qfeat{banktxt}{8974}{}
\end{minipage}\par\nobreak
\vspace{3pt}\nobreak
\noindent{\footnotesize\hangindent=1.4em\hangafter=1 \qnum{1}\enspace  The image \qhl{70}{suggests} that the purpose of the vehicle is to transport passengers and their\qelide{}\par}
\nobreak\vspace{1pt}
\noindent{\footnotesize\hangindent=1.4em\hangafter=1 \qnum{2}\enspace  The scene \qhl{69}{suggests} that someone may be in the process of moving into the cabin,  \qelide{}\par}
\penalty200\vspace{1pt}
\noindent{\footnotesize\hangindent=1.4em\hangafter=1 \qnum{3}\enspace  The image \qhl{69}{suggests} that the man might have some skill or experience in playing frisbee,  \qelide{}\par}
\penalty200\vspace{1pt}
\noindent{\footnotesize\hangindent=1.4em\hangafter=1 \qnum{4}\enspace  The image \qhl{69}{suggests} that the surfers are likely on their way to a beach or  \qelide{}\par}
\penalty200\vspace{1pt}
\par\addvspace{7pt}

\noindent\begin{minipage}{\linewidth}
\qfeat{banktxt}{7634}{}
\end{minipage}\par\nobreak
\vspace{3pt}\nobreak
\noindent{\footnotesize\hangindent=1.4em\hangafter=1 \qnum{1}\enspace  The group of people dressed in suits, who are likely boys, \qhl{70}{seem} \qhl{22}{to} be gathered for a graduation assembly, and they are sitting down\qelide{}\par}
\nobreak\vspace{1pt}
\noindent{\footnotesize\hangindent=1.4em\hangafter=1 \qnum{2}\enspace \qelide{} is someone in the driver's seat of the old truck cabin. They \qhl{68}{seem} \qhl{20}{to} be driving the truck, which is not attached to a trailer.\par}
\penalty200\vspace{1pt}
\noindent{\footnotesize\hangindent=1.4em\hangafter=1 \qnum{3}\enspace  The cat \qhl{67}{seems} \qhl{21}{to} be curiously exploring items in the closet area of the home,\qelide{}\par}
\penalty200\vspace{1pt}
\noindent{\footnotesize\hangindent=1.4em\hangafter=1 \qnum{4}\enspace \qelide{} small plane flying high above the trees. They are gathered together and \qhl{67}{seem} \qhl{20}{to} be enjoying the sight of the aircraft.\par}
\penalty200\vspace{1pt}
\par\addvspace{7pt}

\noindent\begin{minipage}{\linewidth}
\qfeat{banktxt}{8532}{}
\end{minipage}\par\nobreak
\vspace{3pt}\nobreak
\noindent{\footnotesize\hangindent=1.4em\hangafter=1 \qnum{1}\enspace  One giraffe is extending its neck across the field, \qhl{70}{appearing} to interact or look at the other giraffes.\par}
\nobreak\vspace{1pt}
\noindent{\footnotesize\hangindent=1.4em\hangafter=1 \qnum{2}\enspace  The teddy bear is sitting down, \qhl{69}{appearing} to be reading a book, with the book placed in its lap\qelide{}\par}
\penalty200\vspace{1pt}
\noindent{\footnotesize\hangindent=1.4em\hangafter=1 \qnum{3}\enspace  The calico cat is lying down in the grass, \qhl{68}{appearing} stealthy \qhl{19}{and} alert, possibly sneaking up on an insect. It seems to\qelide{}\par}
\penalty200\vspace{1pt}
\noindent{\footnotesize\hangindent=1.4em\hangafter=1 \qnum{4}\enspace  The man \qhl{67}{appears} surprised, and he is making faces at the wall of cuckoo clocks.\par}
\penalty200\vspace{1pt}
\par\addvspace{7pt}

\noindent\begin{minipage}{\linewidth}
\qfeat{banktxt}{12038}{}
\end{minipage}\par\nobreak
\vspace{3pt}\nobreak
\noindent{\footnotesize\hangindent=1.4em\hangafter=1 \qnum{1}\enspace \qelide{} a recreational activity together. They are playing with a frisbee in \qhl{20}{what} \qhl{70}{appears} \qhl{25}{to} be a backyard setting. The presence of a dining table in\qelide{}\par}
\nobreak\vspace{1pt}
\noindent{\footnotesize\hangindent=1.4em\hangafter=1 \qnum{2}\enspace  The setting of the image \qhl{69}{appears} \qhl{20}{to} be a restaurant or dining establishment, as evidenced by the presence\qelide{}\par}
\penalty200\vspace{1pt}
\noindent{\footnotesize\hangindent=1.4em\hangafter=1 \qnum{3}\enspace \qelide{} is traveling on a roadway in a city area, next to \qhl{18}{what} \qhl{68}{appears} \qhl{26}{to} be a Victorian-style building.\par}
\penalty200\vspace{1pt}
\noindent{\footnotesize\hangindent=1.4em\hangafter=1 \qnum{4}\enspace  The setting \qhl{67}{appears} to be a gathering or a party where people are enjoying drinks.\qelide{}\par}
\penalty200\vspace{1pt}
\par\addvspace{7pt}

\noindent\begin{minipage}{\linewidth}
\qfeat{banktxt}{7928}{}
\end{minipage}\par\nobreak
\vspace{3pt}\nobreak
\noindent{\footnotesize\hangindent=1.4em\hangafter=1 \qnum{1}\enspace \qelide{} one notable feature is the presence of many bars on it. This \qhl{70}{may} suggest the use of architectural elements to create an interesting visual effect\qelide{}\par}
\nobreak\vspace{1pt}
\noindent{\footnotesize\hangindent=1.4em\hangafter=1 \qnum{2}\enspace  Yes, there is a tag in the display, which \qhl{69}{might} contain information about the floral arrangements, their creator, or their inspiration. The\qelide{}\par}
\penalty200\vspace{1pt}
\noindent{\footnotesize\hangindent=1.4em\hangafter=1 \qnum{3}\enspace \qelide{} as seen from the overhead view in the image. This organized layout \qhl{68}{might} suggest that the area is well-maintained and possibly set up by a\qelide{}\par}
\penalty200\vspace{1pt}
\noindent{\footnotesize\hangindent=1.4em\hangafter=1 \qnum{4}\enspace \qelide{} is set in a big room with a brick wall visible. This \qhl{68}{may} \qhl{24}{be} an indoor space, such as a party hall, conference room, or\qelide{}\par}
\penalty200\vspace{1pt}
\par\addvspace{7pt}

\noindent\begin{minipage}{\linewidth}
\qfeat{banktxt}{8007}{}
\end{minipage}\par\nobreak
\vspace{3pt}\nobreak
\noindent{\footnotesize\hangindent=1.4em\hangafter=1 \qnum{1}\enspace \qelide{} standing next to his luggage:    Travel arrangements: The man \qhl{70}{might} be discussing his travel plans or coordinating travel-related logistics. He \qhl{39}{could} be        \qelide{}\par}
\nobreak\vspace{1pt}
\noindent{\footnotesize\hangindent=1.4em\hangafter=1 \qnum{2}\enspace \qelide{} table. Some possible explanations include:    Wine tasting: The man \qhl{70}{might} be participating in a wine tasting event, assessing and comparing the flavors,        \qelide{}\par}
\penalty200\vspace{1pt}
\noindent{\footnotesize\hangindent=1.4em\hangafter=1 \qnum{3}\enspace    \qelide{} his personal belongings, such as clothes, toiletries, and souvenirs. The cardboard box \qhl{46}{might} hold items that could not fit into the suitcases or require special           \qelide{}\par}
\penalty200\vspace{1pt}
\noindent{\footnotesize\hangindent=1.4em\hangafter=1 \qnum{4}\enspace \qelide{} Some possible reasons include:    Arrival or departure: The man \qhl{67}{might} have just arrived in the city or is about to depart and           \qelide{}\par}
\penalty200\vspace{1pt}
\par\addvspace{7pt}

\noindent\begin{minipage}{\linewidth}
\qfeat{banktxt}{11807}{}
\end{minipage}\par\nobreak
\vspace{3pt}\nobreak
\noindent{\footnotesize\hangindent=1.4em\hangafter=1 \qnum{1}\enspace \qelide{} rock-covered beach, looking out at the ocean while holding an umbrella. They \qhl{70}{could} \qhl{18}{be} using the umbrella for various reasons, such as to shield themselves\qelide{}\par}
\nobreak\vspace{1pt}
\noindent{\footnotesize\hangindent=1.4em\hangafter=1 \qnum{2}\enspace  The woman \qhl{69}{could} be carrying a tray full of carrots in this setting for various\qelide{}\par}
\penalty200\vspace{1pt}
\noindent{\footnotesize\hangindent=1.4em\hangafter=1 \qnum{3}\enspace  The man \qhl{68}{could} be gesturing near the hazard sign to bring attention to it or\qelide{}\par}
\penalty200\vspace{1pt}
\noindent{\footnotesize\hangindent=1.4em\hangafter=1 \qnum{4}\enspace  The young boy \qhl{68}{could} \qhl{24}{be} wearing the oversized shirt for various reasons, such as playing dress-up,\qelide{}\par}
\penalty200\vspace{1pt}
\par\addvspace{7pt}

\noindent\begin{minipage}{\linewidth}
\qfeat{banktxt}{10439}{}
\end{minipage}\par\nobreak
\vspace{3pt}\nobreak
\noindent{\footnotesize\hangindent=1.4em\hangafter=1 \qnum{1}\enspace \qelide{} man's stern facial expression and crossed arms body language in the image \qhl{70}{could} \qhl{39}{indicate} several possible reasons. It is essential to understand that although these              \qelide{}\par}
\nobreak\vspace{1pt}
\noindent{\footnotesize\hangindent=1.4em\hangafter=1 \qnum{2}\enspace \qelide{} where the standing cow presses its head against the lying cow's head, \qhl{60}{could} \qhl{54}{be} \qhl{34}{attributed} to several reasons. Some possibilities include social bonding, grooming, or\qelide{}\par}
\penalty200\vspace{1pt}
\noindent{\footnotesize\hangindent=1.4em\hangafter=1 \qnum{3}\enspace \qelide{} jackets and riding horses on the flattened sandy area, likely a beach, \qhl{58}{could} \qhl{37}{be} a team of mounted police officers patrolling or conducting a special\qelide{}\par}
\penalty200\vspace{1pt}
\noindent{\footnotesize\hangindent=1.4em\hangafter=1 \qnum{4}\enspace  The gathering of bikers, as seen in the image, \qhl{57}{can} \qhl{42}{have} multiple potential reasons. It \qhl{20}{could} \qhl{19}{be} a casual meetup of friends\qelide{}\par}
\penalty200\vspace{1pt}
\par\addvspace{7pt}

\noindent\begin{minipage}{\linewidth}
\qfeat{banktxt}{12063}{}
\end{minipage}\par\nobreak
\vspace{3pt}\nobreak
\noindent{\footnotesize\hangindent=1.4em\hangafter=1 \qnum{1}\enspace  While enjoying their wine, the two men in the image are \qhl{70}{potentially} bonding over a shared interest in watching a football game, as evidenced\qelide{}\par}
\nobreak\vspace{1pt}
\noindent{\footnotesize\hangindent=1.4em\hangafter=1 \qnum{2}\enspace  In a \qhl{35}{typical} home setting, a person reaching under a bed could \qhl{67}{potentially} find a variety of items, such as:    Lost or          \qelide{}\par}
\penalty200\vspace{1pt}
\noindent{\footnotesize\hangindent=1.4em\hangafter=1 \qnum{3}\enspace  The two sandwiches displayed in an appealing manner on a plate could \qhl{66}{potentially} be for a lunch, dinner, or an appetizer at a gathering or\qelide{}\par}
\penalty200\vspace{1pt}
\noindent{\footnotesize\hangindent=1.4em\hangafter=1 \qnum{4}\enspace \qelide{} actively playing a game. Additionally, numerous books are present throughout the room, \qhl{66}{potentially} indicating that the man may also enjoy reading, and he could have\qelide{}\par}
\penalty200\vspace{1pt}
\par\addvspace{7pt}

\clearpage
\section{SAE Feature Evaluation Metrics}
\label{sec:app-metrics}

We provide detailed definitions of the SAE feature-quality metrics used in
Section~\ref{sec:masae-eval}. All metrics are computed on the held-out
evaluation split using an external CLIP encoder
$\phi=(\phi_V,\phi_T)$. For feature $j \in \{1,\ldots,p\}$, let
$\mathcal{V}_j$ and $\mathcal{T}_j$ denote the vision and text samples on which
the feature activates, respectively, and let $a_{s,j}\geq 0$ denote its
activation on sample $s$. We $\ell_2$-normalize the CLIP embeddings so that
their inner products correspond to cosine similarity.

\subsection{Intra- and Inter-Similarity}
\label{app:intra-inter}

We adopt Intra-Similarity and Inter-Similarity from
\citet{shen2026vl}. For each feature $j$, we collect its top-$n_0$ activating
examples from each modality. Let $\mathcal{V}_j^{(n_0)}$ and
$\mathcal{T}_j^{(n_0)}$ denote these sets. We first mean-pool their CLIP embeddings,
\begin{equation}
u_j^V
=
\frac{1}{|\mathcal{V}_j^{(n_0)}|}
\sum_{s\in\mathcal{V}_j^{(n_0)}} \phi_V(s),
\qquad
u_j^T
=
\frac{1}{|\mathcal{T}_j^{(n_0)}|}
\sum_{s\in\mathcal{T}_j^{(n_0)}} \phi_T(s),
\end{equation}
and normalize the pooled representations,
\begin{equation}
\bar u_j^V = \frac{u_j^V}{\|u_j^V\|_2},
\qquad
\bar u_j^T = \frac{u_j^T}{\|u_j^T\|_2}.
\end{equation}
Let $\mathcal{H}$ denote the set of features that are active in both
modalities and let $H=|\mathcal{H}|$. Intra-Similarity measures cross-modal
semantic consistency within the same feature,
\begin{equation}
\mathrm{Sim}_{\mathrm{intra}}
=
\frac{1}{H}
\sum_{j\in\mathcal{H}}
(\bar u_j^V)^\top \bar u_j^T ,
\end{equation}
while Inter-Similarity measures cross-modal similarity across distinct
features,
\begin{equation}
\mathrm{Sim}_{\mathrm{inter}}
=
\frac{1}{H(H-1)}
\sum_{j\in\mathcal{H}}
\sum_{\substack{\ell\in\mathcal{H}\\ \ell\neq j}}
(\bar u_j^V)^\top \bar u_\ell^T .
\end{equation}
Higher Intra-Similarity indicates that the vision and text examples associated
with the same feature are semantically consistent, while lower
Inter-Similarity indicates better separation between different features. In our implementation, we use the top $n_0=50$ activating examples per modality.

\subsection{Multimodal Monosemanticity}
\label{app:mms}

We adopt multimodal monosemanticity (MMS) from
\citet{kaushik2026decomposing}. MMS measures whether a feature activates on
semantically similar examples across modalities, while weighting each pair by
the strength of the feature's activations.

For feature $j$ and modalities $m,n\in\{V,T\}$, let
$\bm{a}_j^{(m)}\in\mathbb{R}^{M}$ and
$\bm{a}_j^{(n)}\in\mathbb{R}^{N}$ contain the non-zero activations of feature
$j$ on its activating samples from the two modalities. We compute the pairwise
CLIP cosine-similarity matrix
\begin{equation}
S_{rq}
=
\phi_m(s_r)^\top \phi_n(s_q),
\end{equation}
and the corresponding co-activation matrix
\begin{equation}
A_{rq}
=
\left|a_{j,r}^{(m)}a_{j,q}^{(n)}\right|.
\end{equation}
For within-modality comparisons, self-pairs are excluded. We normalize the
co-activation weights,
\begin{equation}
\widetilde A_{rq}
=
\frac{A_{rq}}{\sum_{r',q'} A_{r'q'}},
\end{equation}
and define the feature-level score as
\begin{equation}
\mathrm{MMS}_j(m,n)
=
\sum_{r,q}\widetilde A_{rq}S_{rq}.
\end{equation}

Thus, MMS is the co-activation-weighted average semantic similarity of the
examples associated with a feature. If feature $j$ has no activating samples
in one of the two modalities, its cross-modal score is
$\mathrm{MMS}_j(V,T)=0$, so modality-split features are penalized rather than
excluded.

We report the dictionary-level averages
\begin{equation}
\mathrm{MMS}_{v\text{-}t}
=
\frac{1}{p}\sum_{j=1}^{p}\mathrm{MMS}_j(V,T),
\ 
\mathrm{MMS}_{v\text{-}v}
=
\frac{1}{p}\sum_{j=1}^{p}\mathrm{MMS}_j(V,V),
\ 
\mathrm{MMS}_{t\text{-}t}
=
\frac{1}{p}\sum_{j=1}^{p}\mathrm{MMS}_j(T,T).
\end{equation}
MMS$_{v\text{-}t}$ measures cross-modal semantic coherence, while
MMS$_{v\text{-}v}$ and MMS$_{t\text{-}t}$ measure within-modality coherence.
Higher values indicate greater semantic coherence.

\clearpage
\section{Additional MA-SAE Experiments}
\label{app:additional-masae}

\subsection{Cross-Modal Coverage and Matched-Budget Comparison}
\label{app:coverage}

Table~\ref{tab:monosemanticity} reports MMS over the features scored by each method. MMS assigns zero cross-modal score to features that are active in only one modality, whereas for MA-SAE the score is computed over its designated cross-modal bank. Because this bank comprises only 40\% of the full dictionary, this raises the question of whether MA-SAE's advantage reflects a smaller evaluation set rather than higher feature quality. We therefore report three complementary comparisons in Table~\ref{tab:coverage}. \emph{Coverage} is the fraction of the full dictionary that activates in both modalities. \emph{Bimodal MMS} averages MMS$_{v\text{-}t}$ only over features active in both modalities, while \emph{Matched MMS} restricts each method to the same number of highest co-activating features as MA-SAE: 8{,}192 for LLaVA-1.5-7B, 10{,}240 for LLaVA-1.5-13B, and 7{,}168 for InternVL3-8B.

As shown in Table~\ref{tab:coverage}, baseline coverage is already high (87--100\%), so few baseline features are penalized with a zero cross-modal score, while MA-SAE is capped at 40\% by construction. MA-SAE remains best under both the bimodal and matched MMS comparisons across all three backbones. This shows that its advantage is not explained by differences in cross-modal coverage or feature budget, but instead reflects higher-quality cross-modal features.

\begin{table}[h]
\centering
\small
\setlength{\tabcolsep}{12pt}
\renewcommand{\arraystretch}{1.05}

\begin{tabular}{p{0.34\linewidth}rrr}
\toprule
Method & Coverage (\%) & Bimodal MMS $\uparrow$ & Matched MMS $\uparrow$ \\
\midrule

\multicolumn{4}{l}{\textit{LLaVA-1.5-7B}} \\
\quad Vanilla SAE
& 93.1 & 0.1796 & 0.1897 \\
\quad VL-SAE
& 99.8 & 0.1610 & 0.1717 \\
\quad GSAE
& 99.0 & 0.1763 & 0.1914 \\
\quad \textbf{MA-SAE (ours)}
& 39.9 & \textbf{0.2400} & \textbf{0.2386} \\

\midrule
\multicolumn{4}{l}{\textit{LLaVA-1.5-13B}} \\
\quad Vanilla SAE
& 91.3 & 0.1828 & 0.1943 \\
\quad VL-SAE
& 93.3 & 0.1711 & 0.1777 \\
\quad GSAE
& 98.3 & 0.1793 & 0.1949 \\
\quad \textbf{MA-SAE (ours)}
& 39.9 & \textbf{0.2416} & \textbf{0.2407} \\

\midrule
\multicolumn{4}{l}{\textit{InternVL3-8B}} \\
\quad Vanilla SAE
& 99.6 & 0.1864 & 0.2003 \\
\quad VL-SAE
& 87.5 & 0.1698 & 0.1783 \\
\quad GSAE
& 99.4 & 0.1847 & 0.2002 \\
\quad \textbf{MA-SAE (ours)}
& 39.9 & \textbf{0.2402} & \textbf{0.2387} \\
\bottomrule
\end{tabular}

\caption{
Cross-modal coverage, bimodal-feature MMS quality, and matched-budget MMS comparison. MA-SAE remains best when MMS$_{v\text{-}t}$ is averaged only over features active in both modalities and when the number of evaluated features is matched across methods.
}
\label{tab:coverage}
\end{table}
\clearpage
\subsection{Component Ablation}
\label{app:masae-ablation}

We ablate the two components introduced by MA-SAE: modality-partitioned banks and Barlow Twins alignment. Starting from the vanilla SAE, we first add the modality-partitioned banks and then the Barlow Twins alignment objective to obtain the full MA-SAE. Table~\ref{tab:masae-ablation} reports feature-quality metrics across all three MLLMs. Adding modality-partitioned banks improves feature quality over the vanilla SAE, with the strongest performance achieved when combined with Barlow Twins alignment in the full MA-SAE.

\begin{table}[h]
\centering
\small
\setlength{\tabcolsep}{6pt}
\renewcommand{\arraystretch}{1.12}

\begin{tabular}{lccccc}
\toprule
Method
& MMS$_{v\text{-}t}\uparrow$
& MMS$_{v\text{-}v}\uparrow$
& MMS$_{t\text{-}t}\uparrow$
& Intra-Sim.$\uparrow$
& Inter-Sim.$\downarrow$ \\
\midrule

\multicolumn{6}{l}{\textit{LLaVA-1.5-7B}} \\
\quad Vanilla SAE
& 0.1599 & 0.4670 & 0.6180 & 0.2912 & 0.2656 \\
\quad + Modality Banks
& 0.1733 & 0.5033 & 0.6158 & 0.3016 & 0.2780 \\
\quad + Barlow Twins \textbf{(MA-SAE)}
& \textbf{0.2386} & \textbf{0.6509} & \textbf{0.6960}
& \textbf{0.3513} & \textbf{0.2313} \\

\midrule
\multicolumn{6}{l}{\textit{LLaVA-1.5-13B}} \\
\quad Vanilla SAE
& 0.1589 & 0.4586 & 0.6172 & 0.2934 & 0.2638 \\
\quad + Modality Banks
& 0.1726 & 0.4972 & 0.6187 & 0.3022 & 0.2758 \\
\quad + Barlow Twins \textbf{(MA-SAE)}
& \textbf{0.2407} & \textbf{0.6522} & \textbf{0.6982}
& \textbf{0.3526} & \textbf{0.2318} \\

\midrule
\multicolumn{6}{l}{\textit{InternVL3-8B}} \\
\quad Vanilla SAE
& 0.1843 & 0.5319 & 0.6310 & 0.3043 & 0.2705 \\
\quad + Modality Banks
& 0.1829 & 0.5224 & 0.6266 & 0.3103 & 0.2785 \\
\quad + Barlow Twins \textbf{(MA-SAE)}
& \textbf{0.2387} & \textbf{0.6496} & \textbf{0.6950}
& \textbf{0.3509} & \textbf{0.2303} \\

\bottomrule
\end{tabular}

\caption{
Ablation of MA-SAE components across three MLLMs.
}
\label{tab:masae-ablation}
\end{table}

\clearpage
\section{DPO and mDPO Training Details}
\label{appendix:dpo-training}

\textbf{Preference-Pair Construction.}
For the off-policy experiments, each VLFeedback prompt contains between four and eight candidate responses generated by different MLLMs. For each prompt, we randomly sample two candidate responses and score both using the corresponding preference-labeling method. We construct exactly one preference pair per prompt, assigning the higher-scoring response as preferred and the lower-scoring response as rejected. In the case of an exact score tie, the preferred response is selected at random. For the on-policy experiments, we sample four candidate responses per prompt from the current policy at temperature $1$. We then randomly sample two responses from this candidate set and construct one preference pair using the same procedure described above. All compared preference-labeling methods again operate on the same sampled candidate pairs.

\textbf{Training Configuration.}
Table~\ref{tab:dpo-hparams} summarizes the shared training configuration and objective-specific hyperparameters for DPO and mDPO.

\vspace{0.3cm}
\begin{table}[h]
\centering
\small
\renewcommand{\arraystretch}{1.08}
\begin{tabularx}{0.76\textwidth}{@{}Xc@{}}
\toprule
\textbf{Hyperparameter} & \textbf{Value} \\
\midrule

\multicolumn{2}{c}{\textbf{Training Settings}} \\
\midrule
Optimizer & AdamW \\
Learning rate & $1\times10^{-5}$ \\
LR schedule & WarmupDecayLR \\
Warmup steps & 93 \\
Training epochs & 3 \\
Total optimizer steps & 937 \\
Global batch size & 32 \\
Adam $\beta_1$, $\beta_2$ & 0.9, 0.999 \\
Weight decay & 0.01 \\
LoRA rank $r$ & 64 \\
LoRA $\alpha$ & 128 \\
LoRA dropout & 0.05 \\

\midrule
\multicolumn{2}{c}{\textbf{DPO Settings}} \\
\midrule
$\beta$ & 0.1 \\

\midrule
\multicolumn{2}{c}{\textbf{mDPO Settings}} \\
\midrule
Loss terms & Response + image-conditional + anchor \\
Term weights & $1:1:1$ \\
$\beta$ & 0.1 for all terms \\
Negative image & Random crop \\
Crop area & 1\%--20\% of original image \\

\bottomrule
\end{tabularx}

\caption{
DPO and mDPO training hyperparameters. The same training configuration is used across all backbone models and preference-labeling methods.
}
\label{tab:dpo-hparams}
\end{table}

\clearpage
\raggedbottom
\section{Additional Qualitative Visualizations of MA-SAE Grounding}
\label{app:grounding-qualitative}

\begin{figure}[h]
    \centering
    \includegraphics[width=\linewidth]{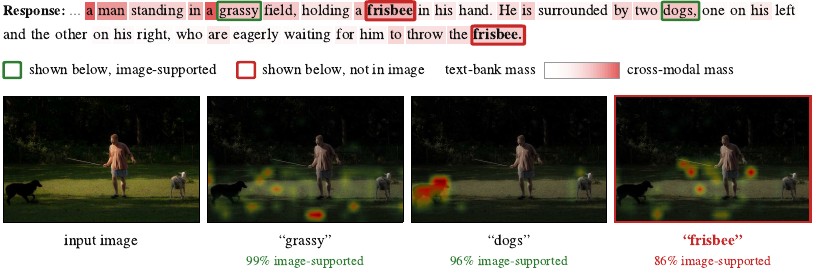}
    \vspace{-0.3cm}
    \caption{
    Additional qualitative visualization of MA-SAE grounding. ``Grassy'' and ``dogs'' receive strong image support, whereas the hallucinated concept ``frisbee'' receives lower support.
    }
    \label{fig:grounding-1}
\end{figure}

\begin{figure}[h]
    \centering
    \includegraphics[width=\linewidth]{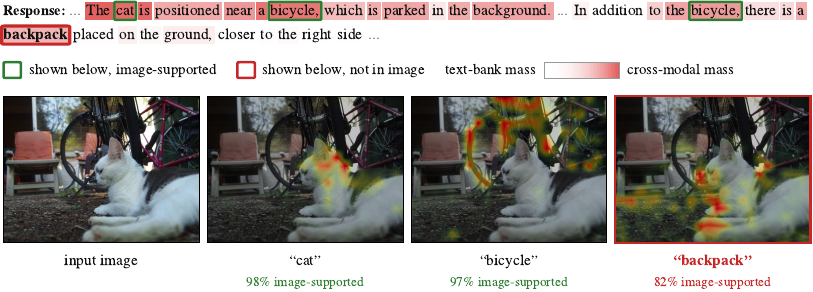}
    \vspace{-0.3cm}
    \caption{
    Additional qualitative visualization of MA-SAE grounding. ``Cat'' and ``bicycle'' receive strong image support, whereas the hallucinated concept ``backpack'' receives lower support.
    }
    \label{fig:grounding-2}
\end{figure}

\begin{figure}[h]
    \centering
    \includegraphics[width=\linewidth]{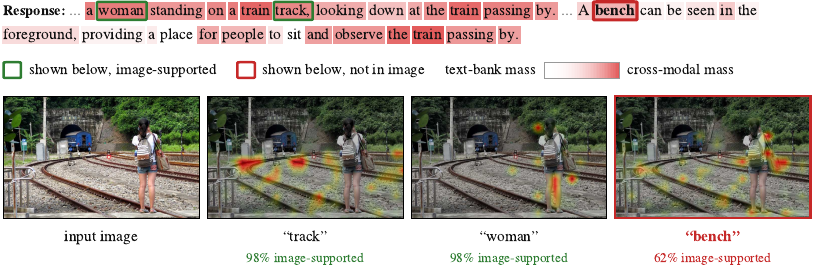}
    \vspace{-0.3cm}
    \caption{
    Additional qualitative visualization of MA-SAE grounding. ``Track'' and ``woman'' receive strong image support, whereas the hallucinated concept ``bench'' receives lower support.
    }
    \label{fig:grounding-3}
\end{figure}

\begin{figure}[H]
    \centering
    \includegraphics[width=\linewidth]{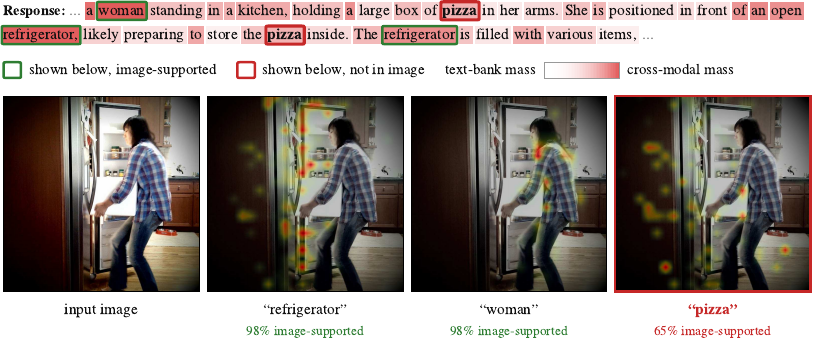}
    \vspace{-0.3cm}
    \caption{
    Additional qualitative visualization of MA-SAE grounding. ``Refrigerator'' and ``woman'' receive strong image support, whereas the hallucinated concept ``pizza'' receives lower support.
    }
    \label{fig:grounding-4}
\end{figure}

\begin{figure}[H]
    \centering
    \includegraphics[width=\linewidth]{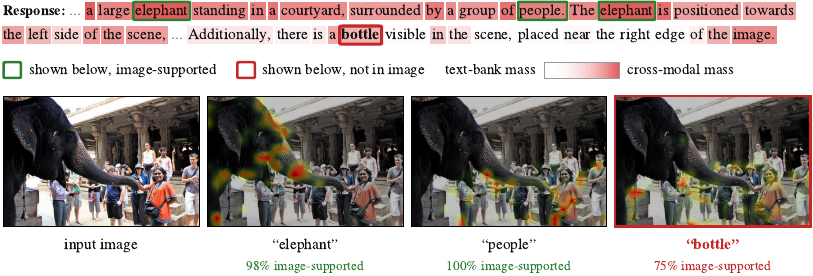}
    \vspace{-0.3cm}
    \caption{
    Additional qualitative visualization of MA-SAE grounding. ``Elephant'' and ``people'' receive strong image support, whereas the hallucinated concept ``bottle'' receives lower support.
    }
    \label{fig:grounding-5}
\end{figure}

Figures~\ref{fig:grounding-1}, \ref{fig:grounding-2}, \ref{fig:grounding-3}, \ref{fig:grounding-4}, and \ref{fig:grounding-5} provide additional qualitative examples of the MA-SAE grounding signal. Across examples, visually supported concepts receive high image support, while hallucinated concepts receive substantially lower support.

\clearpage
\section{Additional Qualitative Comparison of MA-SAE and LLM-as-Judge on AMBER}
\label{app:qualitative-dpo}

\begin{figure}[h]
    \centering
    \includegraphics[width=\linewidth]{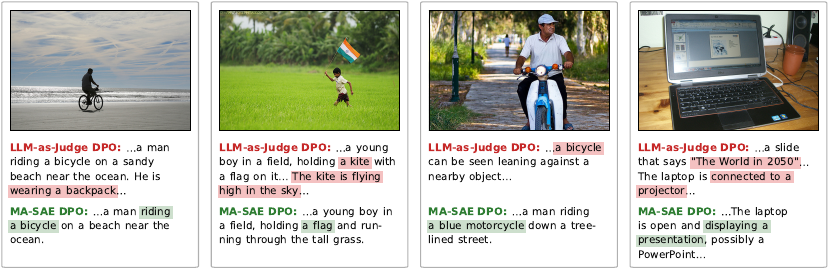}
    \caption{
    Additional AMBER examples comparing LLM-as-Judge DPO and MA-SAE DPO. MA-SAE removes hallucinated objects or attributes and produces more visually grounded responses.
    }
    \label{fig:qualitative-dpo}
\end{figure}

Figure~\ref{fig:qualitative-dpo} provides additional AMBER examples comparing LLM-as-Judge DPO and MA-SAE DPO. Across these examples, MA-SAE removes hallucinated objects or attributes and produces responses that are more closely grounded in the image.

\vspace{2cm}
\section{Ablation of Modality-Specific Banks for Preference Optimization}
\label{app:preference-ablation}

To evaluate whether the modality-specific banks are critical for downstream preference optimization, we compare on-policy DPO performance using the full MA-SAE against a variant that removes the vision- and text-specific banks while retaining the same grounding-score formulation. As shown in Table~\ref{tab:preference-ablation}, removing the modality-specific banks consistently degrades performance across all three MLLM backbones. These results indicate that explicitly isolating modality-specific information is important for constructing a reliable cross-modal grounding signal for preference labeling.

\begin{table}[h]
\centering
\scriptsize
\setlength{\tabcolsep}{3.5pt}
\renewcommand{\arraystretch}{1.05}

\begin{tabular}{lrrrrrrrr}
\toprule
Method
& \multicolumn{4}{c}{AMBER}
& \multicolumn{2}{c}{Object HalBench}
& \multicolumn{2}{c}{MSCOCO} \\
\cmidrule(lr){2-5}
\cmidrule(lr){6-7}
\cmidrule(lr){8-9}

& CHAIR$\downarrow$
& Cover$\uparrow$
& HalRate$\downarrow$
& Cog$\downarrow$
& CHAIR$_s\downarrow$
& CHAIR$_i\downarrow$
& CHAIR$_s\downarrow$
& CHAIR$_i\downarrow$ \\
\midrule

\multicolumn{9}{l}{\textit{LLaVA-1.5-7B}} \\
\quad w/o modality-specific banks
& 6.4 & 50.9 & 28.6 & 2.9
& 42.33 & 11.51
& 50.20 & 16.71 \\
\quad MA-SAE
& \textbf{5.1} & \textbf{51.0} & \textbf{23.8} & \textbf{1.9}
& \textbf{23.00} & \textbf{7.11}
& \textbf{33.40} & \textbf{11.93} \\
\midrule

\multicolumn{9}{l}{\textit{LLaVA-1.5-13B}} \\
\quad w/o modality-specific banks
& 5.4 & 50.7 & 25.4 & 2.4
& 40.33 & 11.40
& 47.80 & 16.43 \\
\quad MA-SAE
& \textbf{4.5} & \textbf{52.4} & \textbf{20.9} & \textbf{1.5}
& \textbf{25.33} & \textbf{6.91}
& \textbf{34.80} & \textbf{11.23} \\
\midrule

\multicolumn{9}{l}{\textit{InternVL3-8B}} \\
\quad w/o modality-specific banks
& 7.3 & \textbf{71.5} & 60.9 & 5.7
& 27.00 & 7.93
& 34.80 & 11.69 \\
\quad MA-SAE
& \textbf{5.9} & 71.1 & \textbf{45.4} & \textbf{3.9}
& \textbf{19.33} & \textbf{5.62}
& \textbf{27.40} & \textbf{9.59} \\
\bottomrule
\end{tabular}

\caption{Ablation of modality-specific banks under on-policy DPO. Removing the vision and text specific banks consistently degrades performance across all three MLLM backbones, demonstrating that explicitly separating modality-specific information is critical for constructing an effective grounding signal.}
\label{tab:preference-ablation}
\end{table}

\clearpage

\section{Analyzing Hallucination Reduction under Matched Response Length and Object Coverage}
\label{app:length-coverage}

\subsection{Off-Policy vs.\ On-Policy Preference-Pair Length Differences}
\label{app:policy-length}

\revision{We first investigate the differences in object coverage between off-policy and on-policy preference tuning. In the off-policy setting, candidate responses are generated by 12 different MLLMs, leading to substantial variation in response length and descriptive breadth. Responses containing fewer unsupported objects often come from less verbose models that produce shorter responses and mention fewer objects overall. Consequently, the model trained with DPO on these preference pairs also learns to generate shorter responses that mention fewer objects, leading to reduced object coverage.}

\revision{In contrast, on-policy candidates are sampled from the same model, resulting in more comparable response lengths. This reduces the confounding effect of response-length differences within preference pairs during DPO training and allows preference tuning to focus more directly on visual grounding.}

\revision{To quantify these differences, we analyze MA-SAE preference pairs for LLaVA-1.5-7B. As shown in Table~\ref{tab:pair-length-analysis}, the median relative length gap within a pair is 52\% in the off-policy setting, compared with only 14\% in the on-policy setting. Moreover, the preferred response is shorter in 76\% of off-policy pairs, compared with 59\% of on-policy pairs. These statistics indicate that on-policy preference pairs are substantially more balanced in response length.}

\begin{table}[h]
\centering
\small
\setlength{\tabcolsep}{8pt}
\renewcommand{\arraystretch}{1.08}
\begin{tabular}{lcc}
\toprule
\revision{MA-SAE preference pairs} & \revision{Off-policy} & \revision{On-policy} \\
\midrule
\revision{Median relative length gap within a pair} & \revision{52\%} & \revision{14\%} \\
\revision{Preferred response is shorter} & \revision{76\%} & \revision{59\%} \\
\bottomrule
\end{tabular}
\caption{\revision{Response-length statistics for MA-SAE preference pairs under off-policy and on-policy tuning with LLaVA-1.5-7B. On-policy preference pairs exhibit substantially smaller length differences.}}
\label{tab:pair-length-analysis}
\end{table}

\revision{These results show that on-policy preference tuning provides a more controlled setting, where preferred and rejected responses have more comparable lengths, mitigating the confounding effect of response-length differences during DPO training. In this setting, MA-SAE achieves both higher object coverage and lower hallucination rates than LLM-as-Judge, demonstrating that its hallucination reductions are not solely attributable to reduced coverage. To further verify that these improvements reflect better visual grounding rather than reduced response verbosity, we conduct two additional analyses: a length-controlled preference-labeling experiment (Section~\ref{app:controlled-preference}) and a matched-coverage evaluation under off-policy tuning (Section~\ref{app:matched-coverage}). Together, these results provide complementary evidence that MA-SAE's hallucination reductions are not merely a consequence of reduced response length or object coverage.}

\subsection{Length-Controlled Preference-Labeling Accuracy}
\label{app:controlled-preference}

\revision{We first directly evaluate whether the MA-SAE grounding score can distinguish factually correct from hallucinated responses independently of response length and linguistic style. We start with over 2,000 hallucination-free captions from MSCOCO and Object HalBench images, verified against ground-truth object annotations. For each caption, we construct a hallucinated counterpart by replacing all mentions of one grounded object with an object absent from the image, using two strategies:}

\begin{itemize}
    \item \revision{\textbf{Random swap:} The replacement object is uniformly sampled from objects absent from the image.}
    \item \revision{\textbf{Plausible swap:} The replacement object is selected from absent objects that frequently co-occur with the scene's objects in COCO (e.g., \emph{fork} $\rightarrow$ \emph{knife} in a dining scene), creating more realistic and challenging hallucinations.}
\end{itemize}

\revision{This controlled setup ensures that each response pair differs only by the substitution of a grounded object with a hallucinated one, thereby preserving linguistic style and response length. We evaluate pairwise preference-labeling accuracy, measured as the proportion of pairs in which the hallucination-free response is preferred.}

\revision{As shown in Table~\ref{tab:controlled-preference}, MA-SAE consistently outperforms LLM-as-Judge across both settings and all three backbones. Notably, MA-SAE maintains high accuracy on plausible swaps, where hallucinated objects are semantically plausible given the scene context. These findings demonstrate that MA-SAE's preference-labeling advantage reflects better visual grounding rather than reduced response verbosity, even when linguistic style and response length are controlled.}

\begin{table}[h]
\centering
\small
\setlength{\tabcolsep}{8pt}
\renewcommand{\arraystretch}{1.08}
\begin{tabular}{lcc}
\toprule
\revision{Method}
& \revision{Random Swap Accuracy (\%)}
& \revision{Plausible Swap Accuracy (\%)} \\
\midrule
\revision{Length} & \revision{50.0} & \revision{50.0} \\
\revision{LLM-as-Judge} & \revision{81.6} & \revision{79.3} \\
\revision{MA-SAE (LLaVA-1.5-7B)} & \revision{\textbf{94.3}} & \revision{\textbf{88.4}} \\
\revision{MA-SAE (LLaVA-1.5-13B)} & \revision{\textbf{96.6}} & \revision{\textbf{90.5}} \\
\revision{MA-SAE (InternVL3-8B)} & \revision{\textbf{95.6}} & \revision{\textbf{88.7}} \\
\bottomrule
\end{tabular}
\caption{\revision{Pairwise preference-labeling accuracy on response pairs controlled for length and linguistic style. MA-SAE consistently outperforms LLM-as-Judge under both settings.}}
\label{tab:controlled-preference}
\end{table}

\subsection{Matched-Coverage Evaluation under Off-Policy Tuning}
\label{app:matched-coverage}

\revision{We next conduct a matched-coverage evaluation under off-policy DPO to investigate whether MA-SAE achieves lower hallucination rates independently of object coverage. Specifically, we compare MA-SAE and LLM-as-Judge on AMBER images for which both models identify exactly the same set of ground-truth objects, ensuring matched coverage. This yields 442 images for LLaVA-1.5-7B and 479 images for LLaVA-1.5-13B.}

\revision{As shown in Table~\ref{tab:matched-coverage}, MA-SAE consistently achieves lower hallucination rates at matched coverage. For LLaVA-1.5-7B, MA-SAE reduces CHAIR from 5.6 to 3.8, HalRate from 31.7 to 15.8, and Cog from 2.4 to 1.5. Similarly, for LLaVA-1.5-13B, CHAIR decreases from 6.1 to 3.9, HalRate from 32.4 to 17.5, and Cog from 2.5 to 1.4. The reductions are statistically significant ($p < 0.001$) and comparable to those observed on the full benchmark, demonstrating that MA-SAE maintains its hallucination reduction advantage over LLM-as-Judge even when object coverage is matched.}

\begin{table}[h]
\centering
\small
\setlength{\tabcolsep}{7pt}
\renewcommand{\arraystretch}{1.08}
\begin{tabular}{lcccc}
\toprule
\revision{Method} & \revision{CHAIR $\downarrow$} & \revision{Cover $\uparrow$} & \revision{HalRate $\downarrow$} & \revision{Cog $\downarrow$} \\
\midrule
\multicolumn{5}{l}{\revision{\textit{LLaVA-1.5-7B}}} \\
\quad \revision{LLM-as-Judge} & \revision{5.6} & \revision{54.7} & \revision{31.7} & \revision{2.4} \\
\quad \revision{MA-SAE} & \revision{\textbf{3.8}} & \revision{54.7} & \revision{\textbf{15.8}} & \revision{\textbf{1.5}} \\
\midrule
\multicolumn{5}{l}{\revision{\textit{LLaVA-1.5-13B}}} \\
\quad \revision{LLM-as-Judge} & \revision{6.1} & \revision{56.1} & \revision{32.4} & \revision{2.5} \\
\quad \revision{MA-SAE} & \revision{\textbf{3.9}} & \revision{56.1} & \revision{\textbf{17.5}} & \revision{\textbf{1.4}} \\
\bottomrule
\end{tabular}
\caption{\revision{Matched-coverage evaluation on AMBER under off-policy DPO, on the images where both models describe exactly the same set of ground-truth objects. At identical coverage, MA-SAE achieves lower CHAIR, HalRate and Cog (CHAIR and HalRate: $p < 0.001$).}}
\label{tab:matched-coverage}
\end{table}

\end{document}